\documentclass[runningheads]{llncs}
\usepackage{graphicx}
\usepackage{amsmath,amssymb} 
\usepackage{booktabs}
\usepackage[breaklinks,colorlinks]{hyperref}
\usepackage[capitalize]{cleveref}
\crefname{section}{Sec.}{Secs.}
\Crefname{section}{Section}{Sections}
\Crefname{table}{Table}{Tables}
\crefname{table}{Tab.}{Tabs.}

\usepackage{newfloat}
\usepackage{listings}
\usepackage{amsmath}
\usepackage{algorithm}

\begin{document}
\title{IoU-Enhanced Attention for End-to-End Task Specific Object Detection}
% \author{Jing Zhao\inst{1} \and
% Shengjian Wu\inst{1} \and
% Li Sun\inst{1, 2}\thanks{Corresponding author, email: sunli@ee.ecnu.edu.cn. \\This work is supported by the Science and Technology Commission of Shanghai Municipality under Grant No. 22511105800, 19511120800 and 22DZ2229004.} \and
% Qingli Li\inst{1}}
%
\authorrunning{J. Zhao et al.}
% \institute{Shanghai Key Laboratory of Multidimensional Information Processing \and
% Key Laboratory of Advanced Theory and Application in Statistics and Data Science\\
% East China Normal University, Shanghai, China}

\author{$\text{Jing Zhao}^1$, $\text{Shengjian Wu}^1$, $\text{Li Sun}^{1,2*}$, $\text{Qingli Li}^1$}
\renewcommand{\thefootnote}{*} 
\footnotetext[1]{Corresponding author, email: sunli@ee.ecnu.edu.cn. \\This work is supported by the Science and Technology Commission of Shanghai Municipality under Grant No. 22511105800, 19511120800 and 22DZ2229004.}
\institute{$^1$Shanghai Key Laboratory of Multidimensional Information Processing, \\
$^2$Key Laboratory of Advanced Theory and Application in Statistics and Data Science,\\ 
East China Normal University, Shanghai, China}
\maketitle              % typeset the header of the contribution
\begin{abstract}
Without densely tiled anchor boxes or grid points in the image, sparse R-CNN achieves promising results through a set of object queries and proposal boxes updated in the cascaded training manner.  
However, due to the sparse nature and the one-to-one relation between the query and its attending region, it heavily depends on the self attention,  
which is usually inaccurate in the early training stage. 
Moreover, in a scene of dense objects, 
the object query interacts with many irrelevant ones, reducing its uniqueness and harming the performance. 
This paper proposes to use IoU between different boxes as a prior for the value routing in self attention.  
The original attention matrix multiplies the same size matrix computed from the IoU of proposal boxes, and they determine the routing scheme so that the irrelevant features can be suppressed.  
Furthermore, to accurately extract features for both classification and regression, we add two lightweight projection heads to provide the dynamic channel masks based on object query, and they multiply with the output from dynamic convs, 
making the results suitable for the two different tasks. 
We validate the proposed scheme on different datasets, including MS-COCO and CrowdHuman, showing that it significantly improves the performance and increases the model convergence speed. 
Codes are available at \href{https://github.com/bravezzzzzz/IoU-Enhanced-Attention}{https://github.com/bravezzzzzz/IoU-Enhanced-Attention}.

% \keywords{First keyword  \and Second keyword \and Another keyword.}
\end{abstract}
\section{Introduction}
\label{sec:intro}

Object detection is a fundamental task in computer vision, which aims to locate and categorize semantic regions with bounding boxes. Traditionally, there are two-stage \cite{girshick2015fast,hu2018relation,ren2015faster} methods based on the densely tiled anchors or one-stage \cite{lin2017focal,liu2016ssd,redmon2016you,redmon2017yolo9000,tian2019fcos} methods built on either anchors or grid points. However, they are both complained for handcrafted designs, \emph{e.g.} the anchor shapes, the standard for positive and negative training samples assignment, and the extra post processing step, like Non-Maximum Suppression (NMS). 

DETR \cite{carion2020end} is a simpler model that utilizes the structure of transformer. In its decoder, a sparse set of learnable queries absorb object contents from the image by cross attention and also from each other by self attention.
Then bipartite matching connected the updated queries and ground truths, assigning only one query to a ground truth box for the classification and regression loss. 
The one-to-one label assignment rule prevents redundant boxes output from the model. 
Therefore, NMS is no longer needed. However, DETR still suffers from the slow training convergence. 
TSP \cite{sun2020rethinking} selects features from the encoder to initialize the query set, and SMCA \cite{gao2021fast} directly predicts a spatial mask to weight the attention matrix. Both of them can accelerate the training speed. 

\begin{figure}[t]
\centering
\includegraphics[width=0.9\columnwidth]{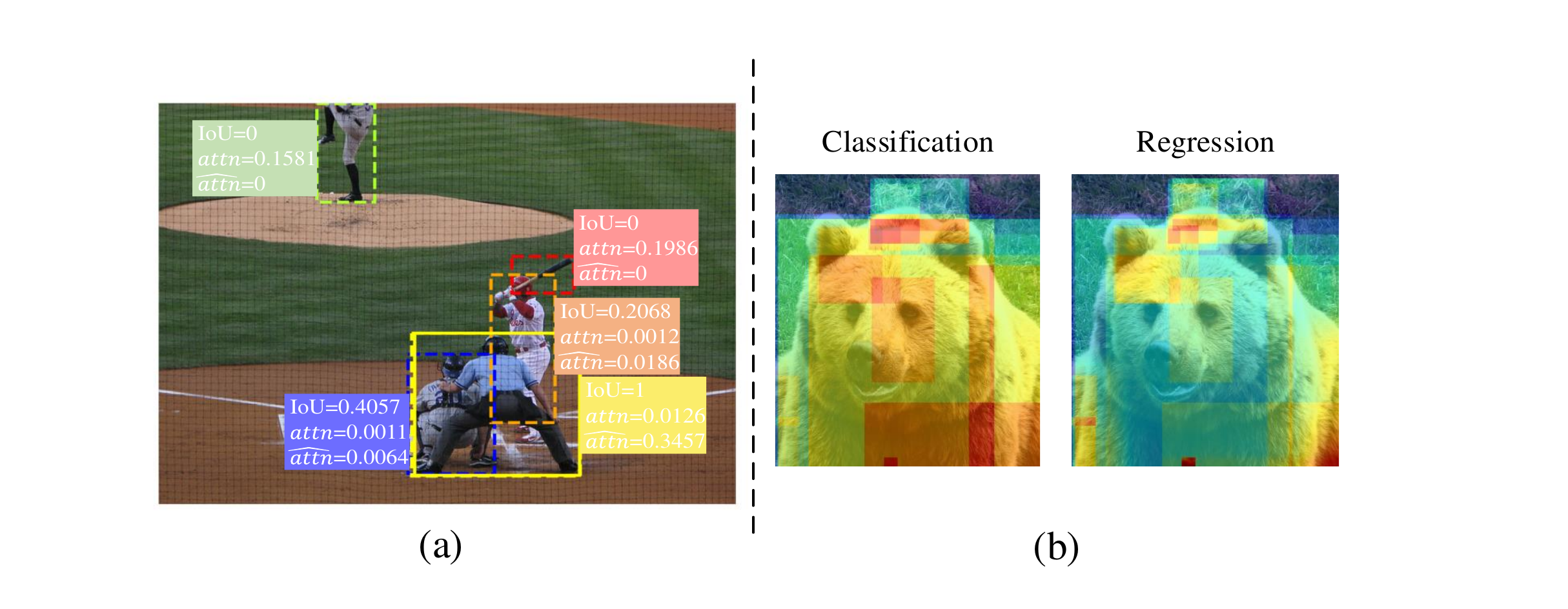}
\caption{Two ways to enhance sparse R-CNN.
(a) focuses on a particular proposal box (in yellow) and its corresponding attention matrix in sparse R-CNN ($attn$) and our model ($\widehat{attn}$). It shows that IoU-ESA can effectively restrict the query attending region, hence keeping the uniqueness of object query.
In (b), two channel masks, dynamically predicted from object queries, 
help to highlight different regions on the features for classification and regression tasks.}
\label{fig:IoU-ESA_DCW}
\end{figure}

The above works have a sparse query set but must attend to the complete image densely.  
Deformable DETR \cite{zhu2020deformable} changes the standard attention into a deformable one, constraining each query to attend to a small set of sampling points around the reference point. 
It significantly reduces the calculations of attention patterns and improves convergence speed. Sparse R-CNN \cite{sun2021sparse} utilizes RoI align \cite{he2017mask} to make local object embedding for loss heads. 
It has paired object query $q_i$ and proposal box $b_i$, and  $q_i$ only interacts with the corresponding image feature within box $b_i$ through dynamic convs, where $i=1,2,\cdots,N$ and $N$ is the total number of the query slots and boxes.
Apart from that, it still performs self attention among the query set so that one $q_i$ knows about the others. 
Sparse R-CNN is a cascaded structure with multiple stages, and $b_i$ is refined in each stage to approach the ground truth box progressively.

Despite the simple designs, sparse R-CNN relies on self attention to model the relationship between different object queries.  
Since $q_i$ provides kernels for each $b_i$, self attention indirectly enlarges the receptive field of the dynamic convs within each box $b_i$. 
However,  $q_i$ is summarized into a vector without spatial dimensions, and the self attention ignores the spatial coverage of $b_i$. 
In other words, $q_i$ may be intensively influenced by other $q_j$ simply because they have similar query and key vectors, without considering the locations of $b_i$ and $b_j$. 
For dense object detection in MS-COCO and CrowdHuman datasets, there are many ground truths of the same class within an image.
The dynamic routing in self attention can be ineffective due to the lack of guidance from the spatial prior, especially in the early training epochs. 

In this paper, we intend to enhance the self attention for object queries in sparse R-CNN by considering the geometry relation between two corresponding boxes, as is shown in \cref{fig:IoU-ESA_DCW}a. 
Similar idea is also adopted in Relation Net \cite{hu2018relation}. 
However, we take a different strategy without adding any parameters. 
The common metric of Intersection over Union (IoU) is directly utilized as a natural prior for value routing during self attention. Specifically, a matrix $IoU$ of size $N\times N$ is computed among $N$ proposal boxes, in which the IoU metric between two boxes $b_i$ and $b_j$ determines each element $IoU_{ij}$.
Note that $IoU$ is in the same size with  the attention matrix $attn$, so the element-wise multiplication can combine them, and the result is used for weighting the value matrix to update object queries. 

To further enhance the feature for classification and localization in the dual loss heads, we design a dynamic channel weighting (DCW) module to produce different and suitable features for two tasks. 
The idea is to add a connection between object query and RoI feature to achieve dynamic cross weighting. It employs the object query $q_i$ to predict two channel masks with the exact dimension of $q_i$. 
Then the two masks are applied to weight the output feature given by the dynamic convs.  
As a result, channels can be highlighted differently for the two tasks, focusing on the relevant regions for classification and regression, as shown in \cref{fig:IoU-ESA_DCW}b. 
Similar to the dynamic conv, there are two light-weight heads in the DCW module, and both have the sigmoid function at the end, like the SE-Net \cite{hu2018squeeze}, constraining the value range from 0 to 1. After the DCW, features for the two tasks are no longer shared, potentially benefiting both. 

The summary of our contributions is listed as follows:
\begin{itemize}
\item We propose an IoU-enhanced self attention module in sparse R-CNN, which brings in the spatial relation prior to the attention matrix. 
Here the IoU of the two proposal boxes is utilized as a metric to evaluate the similarities of different object queries and to guide the value routing in self attention. 
\item We design an inserted DCW module with the cross weighting scheme, which outputs two types of channel masks based on object queries, and then uses them to weight the feature from the original dynamic conv. The DCW separates the RoI features for the classification and regression heads.   
\item Extensive experiments on two datasets of CrowdHuman and MS-COCO are carried out, which validate the proposed method effectively boosts the quality of detection results without significantly increasing the model size and the amount of the calculations.
\end{itemize}

%------------------------------------------------------------------------- 
\section{Related Work}\label{sec:rel}

Object detection \cite{dalal2005histograms,viola2001rapid} has been intensively investigated in computer vision, particularly in the framework of a deep neural network \cite{zhao2019object}. 
Recently, due to the great success of transformer \cite{vaswani2017attention} and its application on image classification \cite{dosovitskiy2020image,liu2021swin,wang2021pyramid}, the performance of detection has been significantly improved due to strong backbones pre-trained on Imagenet \cite{deng2009imagenet}. 
Here we only give a brief review on models, particularly for object detection, primarily including two types which are dense and sparse detectors. 

\subsection{Dense Detector}
Typical dense methods are either two-stage or one-stage. 
From R-CNN \cite{girshick2014rich}, Fast R-CNN \cite{girshick2015fast} to Faster R-CNN \cite{ren2015faster}, two-stage \cite{hu2018relation,song2020revisiting} or even multi-stage \cite{cai2018cascade} methods become tightly fitted into CNN, achieving the end-to-end training manner. 
In these methods, object proposals, which are the roughly located object regions, are first obtained. 
Then, RoI align \cite{he2017mask} extracts these proposal regions from the entire image, making the second stage only observes the RoIs and focuses on improving their qualities. 
One-stage methods detect objects in a single shot without RoI align. 
SSD \cite{liu2016ssd} and YOLO \cite{redmon2016you,redmon2017yolo9000} mimic the first stage in Faster R-CNN, and try to give the final locations and the belonging classes directly . 
However, the detection quality of them significantly lags behind their two-stage competitors. 
RetinaNet \cite{lin2017focal} attributes the inferiority of one-stage methods to the extreme imbalance between the positive and negative training samples and designs a focal loss to deal with it, which down weights the easy negative during training. 
Based on feature pyramid network (FPN) \cite{lin2017feature} and sample assigning schemes, \emph{e.g.,} ATSS \cite{zhang2020bridging}, PAA \cite{kim2020probabilistic} or OTA \cite{ge2021ota}, one-stage method RetinaNet achieves competitive performance. 

Except for the early versions of YOLO \cite{redmon2016you,redmon2017yolo9000}, most of the above works have densely tiled initial boxes in the image, known as anchors. 
However, their sizes and ratios are important but challenging to choose. 
Recent works demonstrate that single-stage anchor-free methods achieve promising results. 
Without tiled anchors, these methods directly classify each grid point and regress bounding boxes for positive samples. 
Different strategies have been adopted to locate a box from a grid. 
FCOS \cite{tian2019fcos} outputs distances to four borders of the ground truth box.
At the same time, it also predicts centerness as a supplement quality metric for NMS.
RepPoints \cite{yang2019reppoints} utilizes deformable conv \cite{dai2017deformable} to spread irregular points around the object center, and then collects all point positions for locating the ground truth box. 
CornerNet \cite{law2018cornernet} adopts a bottom-up scheme, which estimates Gaussian heatmaps for the top-left and bottom-right corners of a ground truth, and tries to match them from the same ground truth through an embedding feature. 
A similar idea is also used to predict heatmaps for object centers \cite{zhou2019objects}. 
RepPoints-v2 \cite{chen2020reppoints} integrates the idea of CornerNet by adding a branch to predict corner heatmaps, which also helps refine the predicted box from irregular points during inference.

\subsection{Sparse Detector}
Although anchor-free methods greatly alleviate the complexity, they still need to predict on the densely tiled grids.
Since a ground truth is potentially assigned to many candidates as positive, NMS becomes the inevitable post processing step to reduce the redundant predictions.
However, NMS is complained for its heuristic design, and many works try to improve \cite{bodla2017soft,huang2020nms,liu2019adaptive} or even eliminate it \cite{sun2020onenet}.
But these works often bring in complex modules. 
DETR \cite{carion2020end} gives a simple NMS-free solution under the transformer structure \cite{vaswani2017attention}. 
It assumes a sparse set of learnable queries updated repeatedly by both self and cross attentions.
The query is finally employed to predict its matching ground truth. 
Note that one ground truth box is only assigned to its belonging query through bipartite matching. Therefore NMS is not required anymore. 
Despite its simple setting, DETR takes a long time to converge. 
TSP \cite{sun2020rethinking} finds that the unstable behavior of cross attention in the early training epochs is the leading cause of the slow convergence. 
Hence it removes cross attention and only performs self attention among selected RoIs. 
Using FPN as backbone, TSP-FCOS and TSP-RCNN can achieve satisfactory results with 36 epochs. 
SMCA \cite{gao2021fast} incorporates the spatial prior predicted directly from query to modulate the attention matrix and increases convergence speed.
Deformable DETR \cite{zhu2020deformable} uses the idea in deformable conv-v2 \cite{zhu2019deformable} into transformer's encoder and decoder. 
Different from traditional non-local attention operation, each query in deformable attention straightly samples the features at irregular positions around the reference point and predicts the dynamic weight matrix to combine them.
The weighted feature updates to query for the next stage until it feeds into the loss heads. 
Sparse R-CNN \cite{sun2021sparse} can be regarded as a variant of DETR and assumes an even simpler setting.  
It has a set of learnable object queries and proposal boxes, and kernels given by query slots process only regions within proposal boxes.
Details are provided in \cref{sec:preliminaries}.
We believe that sparse R-CNN can be easily enhanced without breaking its full sparse assumption and increasing many calculations and parameters.

%------------------------------------------------------------------------- 
\section{Method}\label{sec:method}
\begin{figure}[t]
    \centering
    \includegraphics[width=0.7\columnwidth]{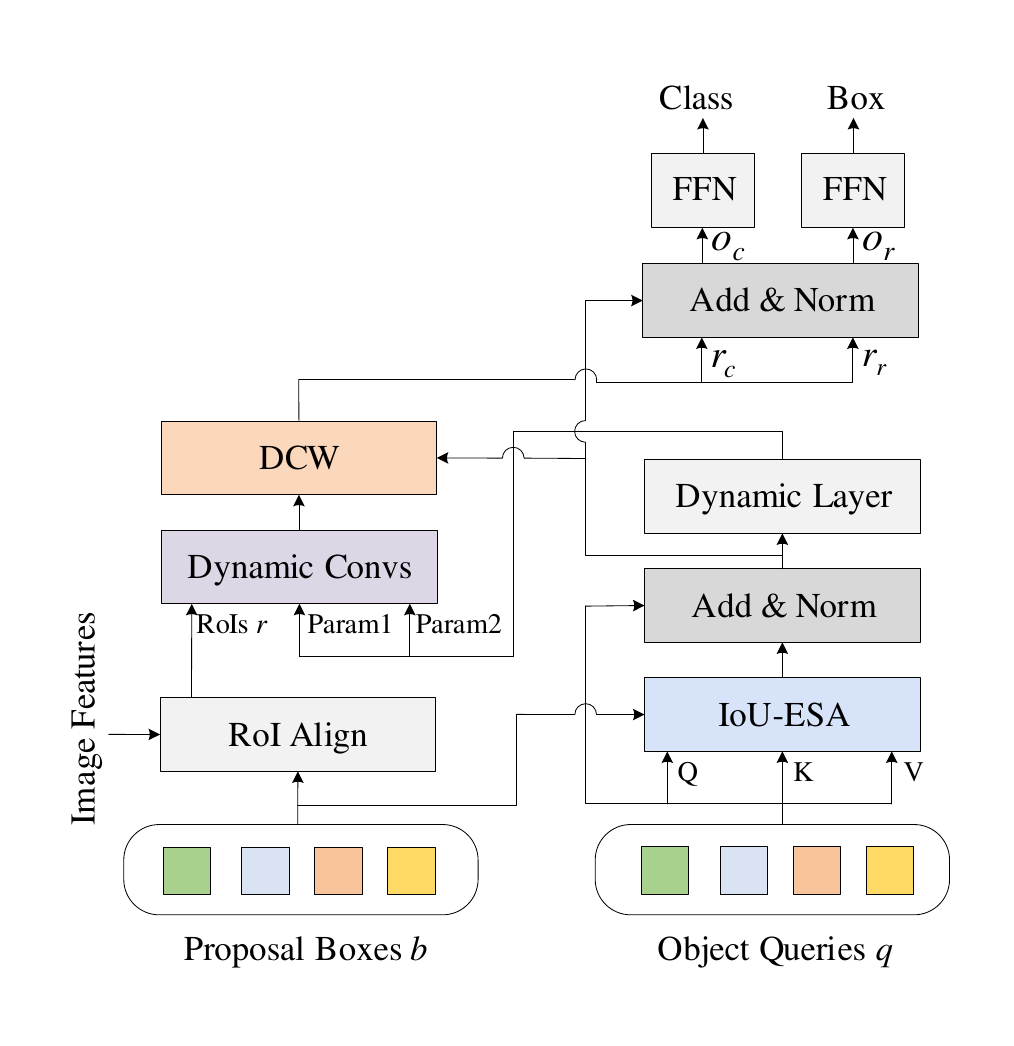}
    \caption{An overview illustration of our approach. 
    Besides the original dynamic convs module in sparse R-CNN, two inserted modules, IoU-Enhanced Self Attention (IoU-ESA) and Dynamic Channel Weighting (DCW), are designed.
    The former enhances the self attention among object queries by IoU between proposal boxes. 
    The latter outputs disentangled features for the two tasks, classification and regression, in object detection. 
    }
    \label{fig:overview}
\end{figure}

We now introduce the proposed method, 
mainly including IoU-enhanced self attention (IoU-ESA) and dynamic channel weighting (DCW). The overview of the method is shown in \cref{fig:overview}. Before illustrating the details of the designed modules, we first review and analyze the preliminary works of sparse R-CNN.

\subsection{Preliminaries and Analysis on Sparse R-CNN}\label{sec:preliminaries}
\subsubsection{Basic formulation.} 
Details about sparse RCNN can also be found in \cref{fig:overview}. 
It has a set of the learnable object queries $q$ to provide one-to-one dynamic 
interactions within their corresponding learnable proposal boxes $b$. 
These $q$ are expected to encode the content of the ground truth boxes. 
The proposal boxes $b$ represent the rectangles.
They locate on features of the whole image for regional RoIs $r$ by RoI align.
Dynamic convs, with the two sets of parameters (Params1-2) generated from the object queries $q$ by a linear projection, is performed on RoI features $r$. 
Then, object features $o$ for dual loss heads can be obtained. Note that the dynamic convs directly link the object queries $q$ with their corresponding RoIs $r$.  
Sparse R-CNN is in an iterative structure with multiple stages. 
Particularly, both $q$ and $b$ are updated in the intermediate stage and potentially used for the next one. 
In the initial stage, $q$ and $b$ are model parameters which can be tuned by back-propagation during training. 

Sparse R-CNN is similar to DETR in the following aspects:
(i) They both have learnable object queries, which intends to describe and capture the ground truths.  
In sparse R-CNN, it is the proposal feature, while in DETR, it is the positional encoding. 
(ii) In these two methods, self attention is performed among object queries to aggregate information from the entire set. 
(iii) Dynamic convs and cross attention are comparable. 
Both perform interactions between object queries and image feature, refining the object embeddings used for classification and regression. 
In summary, sparse R-CNN can be regarded as a variant of DETR depending on RoI align, with a different way to generate object embeddings for loss heads.

\subsubsection{Analysis on self attention.}\label{sec:Analysis_on_self_attention}
The unique property of sparse R-CNN is its sparse-in sparse-out paradigm throughout the processing steps.
Neither dense candidates nor interacting with global features exists in the pipeline. However, we find that the self attention plays an important role, since it models the relation among $q$ and enlarges the receptive field of it. 
Although $q$ is a summarized vector without spatial coverage, the geometry relation is also expected to be discovered in the self attention and to be considered in value routing. 
To validate our assumption, we train one model without self attention and the other in which the self attention matrix is replaced by the IoU matrix computed directly between proposal boxes. Their performances are compared with the original sparse R-CNN in \cref{table:analyze_on_selfattn}. 
These experiments are conducted on MS-COCO.
% Note that these experiments are conducted on CrowdHuman. 
% AP and mMR indicate the Average Precision and Log-average Miss Rate, respectively.
It can be seen that the worst performance, with the AP of 38.5 % and mMR of 65.5
, is from the model w/o MSA without any self attention among object queries.
When IoU matrix is used as the attention matrix, the results from IoU-MSA have been greatly recovered. However, the original self attention still achieves the optimal performance. 
Thus, we intend to enhance the self attention in sparse R-CNN by utilizing the IoU as a geometry prior to guide the routing scheme in self attention. 
Note that, all the self attentions are in the form of multi-head.
The analyze result on CrowdHuman will be presented in the supplementary material.

\begin{figure}[t]
    \centering
    \includegraphics[width=0.8\columnwidth]{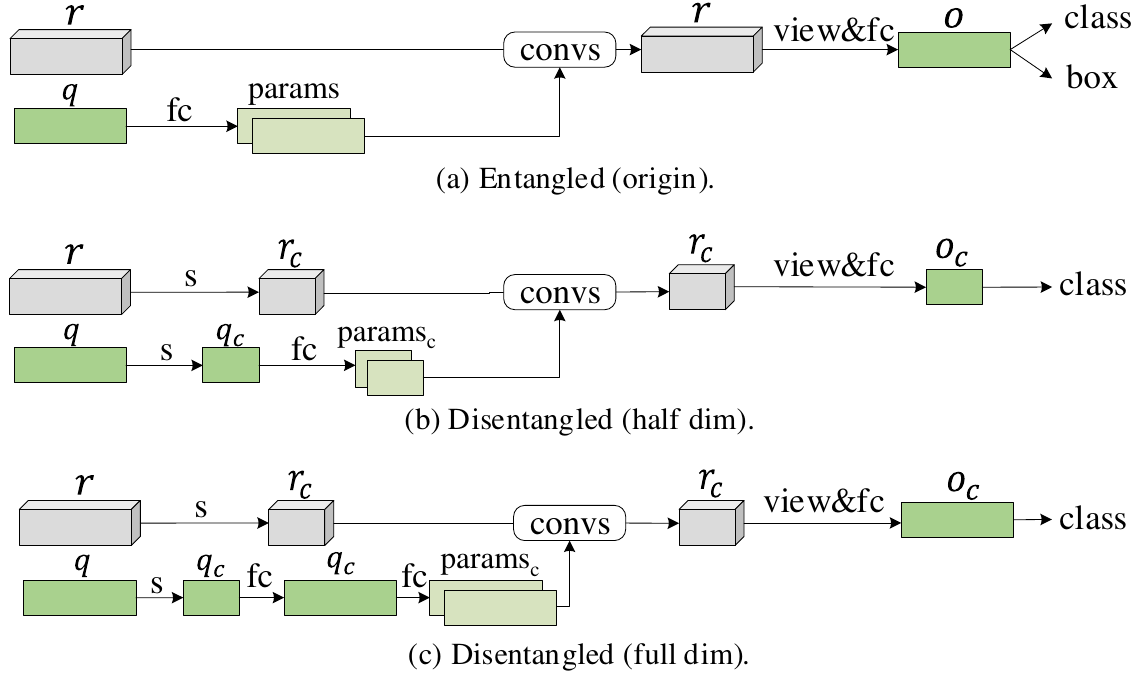}
    \caption{Analysis on feature disentanglement. (a)-(c) indicate the original sparse RCNN, disentangled dynamic conv with half dimension for each branch, and disentangled dynamic conv with a projection head to recover full dimension, respectively. Note that, we only show the classification branch in (b) and (c).
    }
    \label{fig:analyze_dim}
\end{figure}

%------------------------------------------------------------------------- 
\subsubsection{Analysis on feature disentanglement.}\label{sec:Analysis_on_feature_disentanglement}
As is described in \cite{song2020revisiting}, the two tasks in detection concentrate on different parts of objects. 
The classification branch concerns about the region of semantics to support its class type,  
while the regression branch is more related to object contours for accurately locating it.  
Therefore, it is beneficial to disentangle the features for them. 
Note that sparse R-CNN uses the same entangled feature for the two tasks, which can be potentially improved. 

We train two models which have separate features and structures for the classification and regression, as shown in \cref{fig:analyze_dim}. 
Both of them use the same dimension (e.g., 256) setting on object queries $q$ and RoI features $r$.
The first one in \cref{fig:analyze_dim}b allocates half dimensions (e.g., 128) for each task.
Therefore the dynamic conv and its later layers become fully independent.
Note that this setting saves the model parameters and calculations.  
The other in \cref{fig:analyze_dim}c has a similar intention but considers to compensating the dimension lost before giving the feature to dual heads.
The number of channels is recovered to the original dimension as in sparse R-CNN, which is realized by an extra projection layer.
One thing that needs to mention is that although the newly added layer brings along more model parameters and calculations, it is still efficient compared with the original model, mainly due to the significant dimension reduction during dynamic convs. 
Their performances on MS-COCO are listed in \cref{table:analyze_on_feature}. 
Compared to the original sparse R-CNN with entangled features and structures for both tasks, the half dim model metric does not obviously degrade, but the parameters and Flops are apparently less.
The full dim model is slightly better than the original and is still slightly efficient. 
These results clearly demonstrate the effectiveness of task disentanglement in sparse R-CNN. 
However, we think the simple strategy to divide channels for different tasks does not fully consider the dynamic weighting scheme from the object query. 
Based on this observation, we intend to enhance the feature for classification and localization by a dynamic channel weighting (DCW) module, which gives a better disentanglement for the two tasks at the cost of only a slight increase in model parameters. 
\begin{table}[ht]
\parbox{.48\linewidth}{
\centering
\caption{Analysis on self attention. 
Three different models, including origin sparse R-CNN with multi-head self attention (MSA), 
without self attention and attention matrix replaced by IoU matrix (IoU-MSA), are trained on MS-COCO. 
}
\vspace{0.2cm}
\label{table:analyze_on_selfattn}
\setlength{\tabcolsep}{5pt}
\begin{tabular}{@{}cccc@{}}
\toprule
Method & AP & Params &  FLOPs \\ \midrule
MSA & 42.8 & 106.2M & 133.8G \\
w/o MSA & 38.5 & 104.7M & 133.6G \\
IoU-MSA & 41.8 & 105.4M & 133.7G \\ \bottomrule
\end{tabular}}
\hfill
\parbox{.48\linewidth}{
\centering
\caption{Analysis on feature disentanglement on MS-COCO. 
AP, the number of parameters, and the amount of calculations of the three models are listed. `En.' and `Dis.' indicate entangled and disentangled feature dimension.
}\label{table:analyze_on_feature}
\vspace{0.2cm}
\setlength{\tabcolsep}{4pt}
\begin{tabular}{@{}cccc@{}}
\toprule
Method & AP & Params &  FLOPs \\ \midrule
En. (original) & 42.8 & 106.2M & 133.8G \\
Dis. (half dim) & 42.7 & 97.0M & 132.9G \\
Dis. (full dim) & 43.1 & 103.7M & 133.5G \\ \bottomrule
\end{tabular}}
\end{table}
\setlength{\tabcolsep}{1.4pt}

\begin{figure}[ht]
    \centering
    \includegraphics[width=.95\columnwidth]{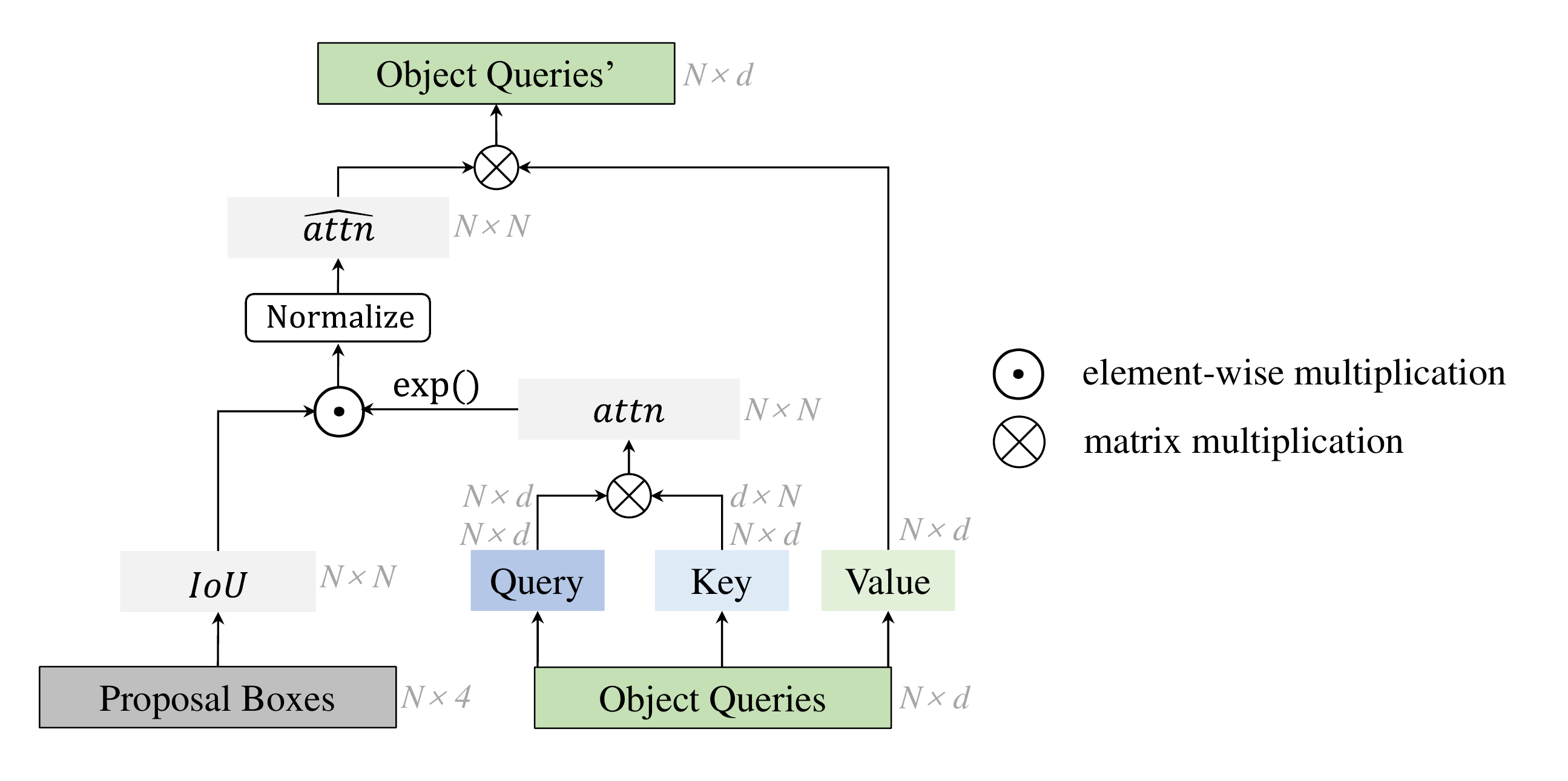}
    \caption{%An illustration 
    Details of 
    IoU-ESA 
    module. Normal self attention is still executed among object queries. The attention matrix $attn$, calculated from Q and K, is further modulated %routed 
    by $IoU$ matrix through element-wise multiplication. $IoU$ matrix reflects similarity 
    of proposal boxes, and it also %together 
    determines the value routing scheme.}
    \label{fig:IoU-ESA}
\end{figure}

%------------------------------------------------------------------------- 
\subsection{IoU-Enhanced Self Attention (IoU-ESA)}
As is depicted in \cref{sec:Analysis_on_self_attention}, 
multi-head self attention (MSA) is applied on the object queries $q$ to obtain the global information from each other. 
Here we further illustrate its mechanism as in \cref{equ.1}, 
\begin{equation}
    {\rm MSA}(\textbf{Q} ,\textbf{K}, \textbf{V}) = [{\rm Attn}\left(\textbf{Q}_i, \textbf{K}_i,\textbf{V}_i\right)]_{i = 1: H}\textbf{W}^o
    \label{equ.1}
\end{equation}
\begin{equation}
    {\rm Attn}\left(\textbf{Q}_i, \textbf{K}_i, \textbf{V}_i\right) = {\rm softmax}\left(\frac{\textbf{Q}_i\textbf{K}_i^{{\rm T}}}{\sqrt{d}}\right)\textbf{V}_i\label{equ.2}
\end{equation}
where $\textbf{Q, K, V} \in{\mathbb{R} ^ {N \times {d}}}$ are the input query, key and value matrices, projected from object queries $q$ by parallel layers with different parameters. 
$d$ is the hidden dimension after the projection, and $H$ is the total number of heads. 
${\rm Attn}\left(\textbf{Q}, \textbf{K}, \textbf{V}\right)$ denotes the standard attention function with \textbf{Q}, \textbf{K}, and \textbf{V} as its input, and it is further clarified in \cref{equ.2}. 
$\textbf{W}^o$ is a set of learnable parameters used to combine multiple heads, and the subscripts specify the index of the head. 
Note that each head independently routes its own value \textbf{V} according to the similarity between \textbf{Q} and \textbf{K}. 

We intend to enhance the self attention by considering the geometry relation between two corresponding boxes, and the details are given in \cref{fig:IoU-ESA}. 
Basically, IoU-ESA utilizes another way to measure the similarity between \textbf{Q} and \textbf{K}. 
Compared with exhaustively measure between any slot pair through $attn=\textbf{Q} \textbf{K}^{\rm  T}/\sqrt{d}\in{\mathbb{R}}^{N \times N}$, IoU is a good prior and it is easier to compute.  
IoU-ESA takes advantage of it without adding any parameters. 
Specifically, each element of $attn$ before $\rm{normalize}$ is multiplied by its corresponding element in $IoU$ matrix, which is computed from the pairs of proposal boxes. 
The IoU-enhanced self attention is formulated as:
\begin{equation}
    \widehat{attn_{ij}} = \frac{{\rm exp}\left(attn_{ij}\right)\cdot
    IoU_{ij}
    }{\sum_{k=1}^{N}{\rm exp}\left(attn_{ik}\right)\cdot
    IoU_{ik}
    }\label{equ.3}
\end{equation}
where $i, j$ are indexed from 1 to $N$, $\widehat{attn}\in{\mathbb{R}}^{N \times N}$ is the enhanced attention matrix, and $IoU \in{\left[0, 1\right]}^{N \times N}$ is measured among $N$ proposal boxes based on the IoU metric. 

\begin{figure}[ht]
    \centering
    \includegraphics[width=0.9\columnwidth]{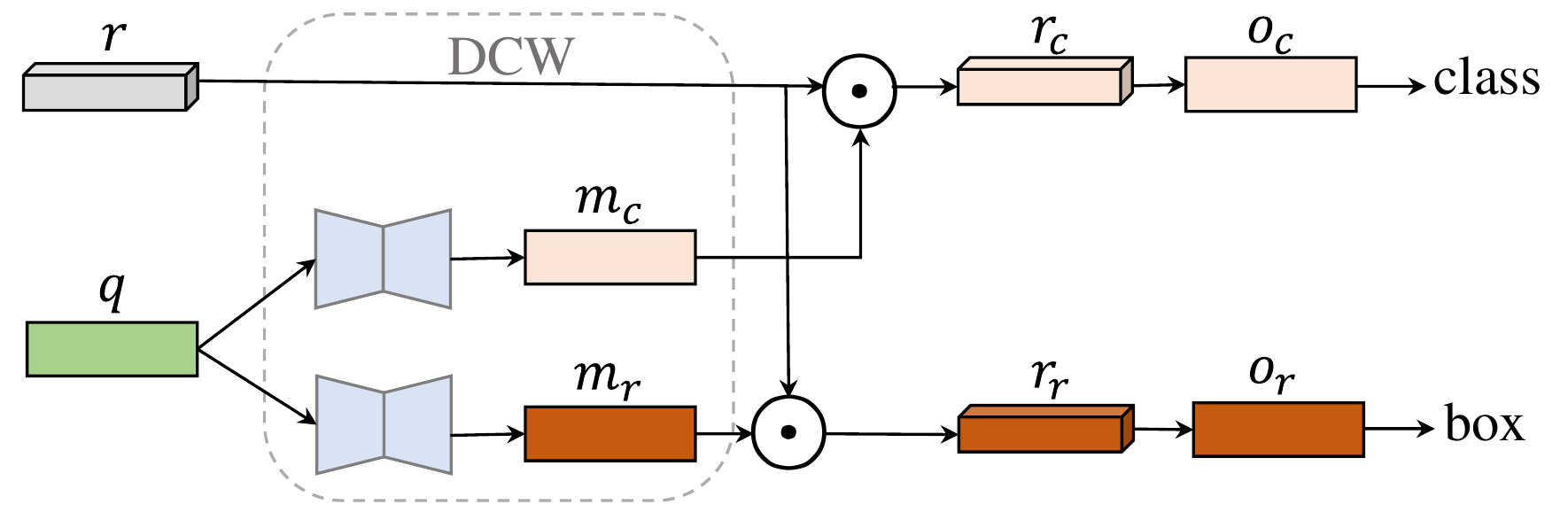}
    \caption{
    Details of 
    DCW module and its way to be applied on RoI feature. %). 
    Two channel masks $m_c$ and $m_r$ are generated from object queries $q$ by the lightweight bottlenecks, and they enhance the RoI features $r$ %out 
    output from dynamic convs. 
    }
    \label{fig:dcw}
\end{figure}

%------------------------------------------------------------------------- 
\subsection{Dynamic Channel Weighting (DCW)}
As is analyzed in \cref{sec:Analysis_on_feature_disentanglement}, it is better to use different features for the two loss heads in detection. 
However, sparse R-CNN sends the same features into both heads, which is obviously not optimal. 
Although the simple strategy, to separate object queries $q$ and RoI features $r$ along the channel dimension, can reduce the parameters and calculations without degrading the performance, a better dynamic scheme, which fully utilizes $q$, is still needed. 

Here we propose a dynamic channel weighting (DCW) module to separate and strengthen the two features. As shown in \cref{fig:dcw}, two channel masks $m_c$ and $m_r\in\mathbb{R}^{N \times d}$ are generated by linear projection layers in parallel with the dynamic layer  for the parameters in dynamic convs.  
The $\rm sigmoid$ is used as the activation, constraining each element in $m_c$ and $m_r$ between 0-1. 
Different from SE-Net, the DCW is actually a dynamic cross weighting scheme. 
Based on the object queries $q$, they generate $m_c$ and $m_r$ with the same number of channels as the RoI features $r$, and the value in them shares for each spatial position in $r$. 
These two channel masks highlight different areas in the proposal box, hence producing more relevant features for classification and regression branches. 
The DCW module can be formulated as follows:
\begin{equation}
    r' = r \odot \sigma\left({\rm g_m}\left(q \right) \right) \label{equ.5}
\end{equation}
where $q$ and $r$ indicate object queries and RoI features after dynamic conv. 
${\rm g_m}$ is a bottleneck including two connected linear projection layers, predicting two masks based on $q$.
$\sigma(\cdot)$ represents the sigmoid activation. 
The result $r'$ is the enhanced RoI features for different tasks, and it can be $r_c$ or $r_r$ as is shown in \cref{fig:dcw}. 
Note that the separate $r_c$ and $r_r$ go through independent layers, which project them into object embeddings $o_c$ and $o_r$. 
These features are finally given to the two loss heads to compute different types of penalties.

Since the original sparse R-CNN is built in cascaded structure, the object queries $q$ need to be updated for the next stage. 
To keep it simple, in our implementation, $o_c$ and $o_r$ are added together to form the new $q$ in the next stage.

%------------------------------------------------------------------------- 
\subsection{Training Objectives}
Following \cite{carion2020end,stewart2016end,sun2021sparse,zhu2020deformable}, we take the Hungarian algorithm to make the prediction and ground truth pairs. 
The bipartite matching manner avoids post processing like NMS. 
The matching cost is the same as the training loss, $\mathcal{L} = \lambda_{cls}\cdot \mathcal{L}_{cls} + \lambda_{L1}\cdot \mathcal{L}_{L1} + \lambda_{giou}\cdot \mathcal{L}_{giou}$. 
Here, $\mathcal{L}_{cls}$ is focal loss \cite{lin2017focal} used for classification.
$\mathcal{L}_{L1}$ and $\mathcal{L}_{giou}$ representing the L1 loss and generalized IoU loss \cite{rezatofighi2019generalized} are employed for box regression. $\lambda_{cls}$, $\lambda_{L1}$ and $\lambda_{giou}$ are coefficients balancing the impact of each term.

%===========================================================
\section{Experiment}
\noindent\textbf{Datasets.}\quad
We perform extensive experiments with variant backbones on the pedestrian detection benchmark CrowdHuman \cite{shao2018crowdhuman} and 80-category MS-COCO 2017 detection dataset \cite{lin2014microsoft}.

For CrowdHuman, training is performed on $\sim$15k train images and testing is evaluated on the $\sim$4k validation images. 
There are a total of 470k human instances from train and validation subsets and 23 persons per image on average, with various kinds of occlusions in the dataset. 
Each human instance is annotated with a head bounding-box, human visible-region bounding-box and human full-body bounding-box. 
We only use the full-body bounding-box for training. 
For evaluation, we follow the standard Caltech \cite{dollar2011pedestrian} evaluation metric mMR, which stands for the Log-Average Missing Rate over false positives per image (FPPI) ranging in $[10^{-2}, 100]$. Average Precision (AP) and Recall are also provided to better evaluate our methods.

For COCO dataset, training is performed on train2017 split ($\sim$118k images) and testing is evaluated on val2017 set (5k images). 
Each image is annotated with bounding-boxes. On average, there are 7 instances per image, up to 63 instances in a single image in the training set, ranging from small to large on the same image.  
If not specified, we report AP as bbox AP, the integral metric over multiple thresholds. We report validation AP at the last training epoch.

~\\
\noindent\textbf{Implementation details.}\quad
All experiments are implemented based on Detectron2 codebase \cite{Detectron2018}.
We initialize backbone from the pre-trained model on ImageNet \cite{deng2009imagenet}, and other newly added layers with Xavier \cite{glorot2010understanding}. 
By default, we treat ResNet-50 \cite{he2016deep} as the backbone, and use FPN \cite{lin2017feature} to deal with different sizes of ground truth boxes.  
AdamW \cite{loshchilov2018decoupled} is employed to optimize the training loss with $10^{-4}$ weight decay.
The initial learning rate is $2.5\times 10^{-5}$, divided by 10 at 40th epoch for CrowdHuman and by 10 at 27th and 33rd epochs for COCO.
All models are trained on 4 RTX 3090 GPUs with batch size of 16, for 50 epochs on CrowdHuman, while 36 epochs on COCO.  
Images are resized such that the shorter edge is at most 800 pixels while the longer side is 1,333. 
The loss weights are set as $\lambda_{cls} = 2$, $\lambda_{L1} = 5$ and $\lambda_{giou} = 2$.
Following sparse R-CNN, the default number of object queries and proposal boxes are 500 for CrowdHuman, 100 for COCO. All models have 6 iterative stages.
The gradients on proposal boxes are detached in the next stage for stable training.
There is no post processing step like NMS at the evaluation.

\subsection{Ablation Study}
\subsubsection{Ablation study on CrowdHuman.}
CrowdHuman consists of images with highly overlapped pedestrians. 
IoU-ESA restricts the attending range of the object queries $q$ during self attention, according to \cref{equ.3}, so $q$ can intentionally consider the geometry relation for updating itself. 
On the other hand, DCW gives the independent feature representations for classification and regression, which is also conducive to improve the performance.
\cref{table:ablation_on_crowdhuman} shows the detailed ablations on them. 
It clearly indicates that both modules can significantly improve the performance. 
When the two modules are combined together, they can bring further enhancements.

\vspace{-0.5cm}
\subsubsection{Ablation study on COCO.}
We further carry out a similar experiment on the richer and larger dataset MS-COCO. 
The results are shown in \cref{table:ablation_on_coco}. Note that compared to CrowdHuman, COCO has variety of classes and dense objects at the same time.
It can be seen that both IoU-ESA and DCW modules can also perform well on COCO.  

\vspace{-0.5cm}
\begin{table}[ht]
\parbox{.48\linewidth}{
\centering
\caption{Ablation study results on CrowdHuman. All experiments are based on ResNet-50, and number of object queries and proposal boxes are 500. }
% \vspace{0.2cm}
\begin{tabular}{@{}ccccc@{}}
\toprule
IoU-ESA & DCW & AP $\uparrow$ &  mMR $\downarrow$ & Recall $\uparrow$ \\ \midrule
%   &   & 88.8 & 49.9 & 95.8 \\
  &   & 89.2 & 48.3 & 95.9 \\
 \checkmark &   & 90.4 & 47.8 & 96.7 \\
  & \checkmark & 89.9 & 47.4 & 96.3 \\
 \checkmark & \checkmark & \textbf{90.9} & \textbf{47.2} & \textbf{96.9} \\\bottomrule
\end{tabular}
\label{table:ablation_on_crowdhuman}}
\hfill
\parbox{.48\linewidth}{
\centering
\caption{Ablation study results on COCO. All experiments are based on ResNet-50, and number of object queries and proposal boxes are set to 100.}
\vspace{0.2cm}
\setlength{\tabcolsep}{5pt}
\begin{tabular}{@{}ccccc@{}}
\toprule
IoU-ESA & DCW & AP &  $\text{AP}_{50}$ & $\text{AP}_{75}$ \\ \midrule
  &   & 42.8 & 61.2 & 45.7 \\
 \checkmark &   & 43.9 & 63.0 & 47.6 \\
  & \checkmark & 43.6 & 62.3 & 47.3 \\
 \checkmark & \checkmark & \textbf{44.4} & \textbf{63.4} & %\textbf{48.2}
 \textbf{48.3} \\\bottomrule
\end{tabular}
\label{table:ablation_on_coco}}

\end{table}
\setlength{\tabcolsep}{1.4pt}
\vspace{-1cm}

\subsection{Main Results}
\subsubsection{Main results on CrowdHuman.}
To thoroughly evaluate the performance of our proposed methods, we do plenty of experiments based on the same backbone (ResNet-50) for different methods. 
The results are shown in \cref{table:result_on_crowdhuman}.
It obviously shows that our work outperforms the sparse R-CNN and other works especially the sparse detector, including DETR and deformable DETR. 
Note that we increase all metrics, including AP, mMR and Recall, at the same time.
Particularly, the enhanced version can achieve +1.7, -1.1, +1.0 gains on AP, mMR and Recall, respectively, compared with sparse R-CNN. 

\setlength{\tabcolsep}{8pt}
\begin{table}[ht]
\centering
\caption{Performance comparisons on CrowdHuman. }
\label{table:result_on_crowdhuman}
\vspace{0.2cm}
\begin{tabular}{c|c|ccc}
\toprule
Method & NMS & AP $\uparrow$ &  mMR $\downarrow$ & Recall $\uparrow$\\ \midrule
 Faster R-CNN & \checkmark & 85.0 & 50.4 & 90.2 \\
 RetinaNet & \checkmark & 81.7 & 57.6 & 88.6 \\
 FCOS & \checkmark & 86.1 & 55.2 & 94.3 \\
 AdaptiveNMS & \checkmark & 84.7 & 49.7 & 91.3 \\
 DETR & $\circ$ & 66.1 & 80.6 & - \\
 Deformable DETR & $\circ$ & 86.7 & 54.0 & 92.5 \\
%   Sparse R-CNN & $\circ$ & 88.8 & 49.9 & 95.8 \\
 Sparse R-CNN & $\circ$ & 89.2 & 48.3 & 95.9 \\ \midrule
 ours & $\circ$ & \textbf{90.9} & \textbf{47.2} & \textbf{96.9} \\
\bottomrule
\end{tabular}
\end{table}
\setlength{\tabcolsep}{1.4pt}
\vspace{0.2cm}

\setlength{\tabcolsep}{1pt}
\begin{table}[b]
\centering
\caption{Performances of different detectors on COCO 2017 val set. Here `*' indicates that the model is with 300 object queries
and random crop training augmentation, `Res' indicates ResNet. The top two sections show results from Detectron2 \cite{Detectron2018}, mmdetection \cite{chen2019mmdetection} 
or original papers \cite{carion2020end,gao2021fast,sun2021sparse,zhu2020deformable}.}
\label{table:res_coco_val}
% \vspace{0.2cm}
\begin{tabular}{c|cccc|cccccc}
\toprule
Method & Backbone & Epoch & Params & FLOPs & $\text{AP}$ &  $\text{AP}_{50}$ & $\text{AP}_{75}$ & $\text{AP}_S$ & $\text{AP}_M$ & $\text{AP}_L$ \\ \midrule
 RetinaNet\cite{Detectron2018}  & Res-50 & 36 & 38M & 206G & 38.7 & 58.0 & 41.5 & 23.3 & 42.3 & 50.3 \\
 RetinaNet\cite{Detectron2018} & Res-101 & 36 & 57M & 273G & 40.4 & 60.3 & 43.2 & 24.0 & 44.3 & 52.2  \\
 Faster R-CNN\cite{Detectron2018} & Res-50 & 36 & 42M & 180G & 40.2 & 61.0 & 43.8 & 24.2 & 43.5 & 52.0 \\
 Faster R-CNN\cite{Detectron2018} & Res-101 & 36 & 61M & 232G & 42.0 & 62.5 & 45.9 & 25.2 & 45.6 & 54.6 \\
 Cascade R-CNN\cite{chen2019mmdetection} & Res-50 & 20 & 69M & 235G & 41.0 & 59.4 & 44.4 & 22.7 & 44.4 & 54.3 \\
%  Cascade R-CNN \cite{chen2019mmdetection} & ResNet-101 & 20 & 311 & 42.5 & 60.7 & 46.4 & 23.5 & 46.5 & 56.4 \\
 DETR\cite{carion2020end} & Res-50 & 500 & 41M & 86G & 42.0 & 62.4 & 44.2 & 20.5 & 45.8 & 61.1 \\
 DETR\cite{carion2020end} & Res-101 & 500 & 60M & 152G & 43.5 & 63.8 & 46.4 & 21.9 & 48.0 & 61.8 \\
%  DETR \cite{carion2020end} & ResNet-50-DC5 & 500 & 187 & 43.3 & 63.1 & 45.9 & 22.5 & 47.3 & 61.1 \\
%  DETR \cite{carion2020end} & ResNet-101-DC5 & 500 & 253 & 44.9 & 64.7 & 47.7 & 23.7 & 49.5 & 62.3 \\
 Deformable DETR\cite{zhu2020deformable} & Res-50 & 50 & 40M & 173G & 43.8 & 62.6 & 47.7 & 26.4 & 47.1 & 58.0 \\
 TSP-FCOS\cite{sun2020rethinking} & Res-50 & 36 & 52M & 189G & 43.1 & 62.3 & 47.0 & 26.6 & 46.8 & 55.9 \\
 TSP-RCNN\cite{sun2020rethinking} & Res-50 & 36 & 64M & 188G &  43.8  & 63.3 & 48.3 & 28.6 & 46.9 & 55.7 \\
 TSP-FCOS\cite{sun2020rethinking} & Res-101 & 36 & - & 255G & 44.4 & 63.8 & 48.2 & 27.7 & 48.6 & 57.3 \\
 TSP-RCNN\cite{sun2020rethinking} & Res-101 & 36 & - & 254G &  44.8 & 63.8 & 49.2 & 29.0 & 47.9 & 57.1  \\
 SMCA\cite{gao2021fast} & Res-50 & 50 & 40M & 152G & 43.7 & 63.6 & 47.2 & 24.2 & 47.0 & 60.4 \\
 SMCA\cite{gao2021fast} & Res-101 & 50 & 58M & 218G & 44.4 & 65.2 & 48.0 & 24.3 & 48.5 & 61.0  \\
 \midrule
 Sparse R-CNN\cite{sun2021sparse} & Res-50 & 36 & 106M & 134G & 42.8 & 61.2 & 45.7 & 26.7 & 44.6 & 57.6 \\
 Sparse R-CNN\cite{sun2021sparse} & Res-101 & 36 & 125M & 201G & 44.1 & 62.1 & 47.2 & 26.1 & 46.3 & 59.7 \\
 Sparse R-CNN*\cite{sun2021sparse} & Res-50 & 36 & 106M & 152G & 45.0 & 63.4 & 48.2 & 26.9 & 47.2 & 59.5 \\
 Sparse R-CNN*\cite{sun2021sparse}& Res-101 & 36 & 125M & 219G & 46.4 & 64.6 & 49.5 & 28.3 & 48.3 & 61.6 \\
 Sparse R-CNN*\cite{sun2021sparse}& Swin-T & 36 & 110M & 158G & 47.9 & 67.3 & 52.3 & - & - & - \\\midrule
%  ours & Res-50 & 36 & 133M & 137G & 44.4 & 63.4 & 48.2 & 27.5 & 47.2 & 58.2 \\
%  ours & Res-101 & 36 & 152M & 203G & 45.6 & 64.4 & 49.8 & 27.5 & 48.5 & 60.2 \\
%  ours* & Res-50 & 36 & 133M & 160G & 46.5 & 65.9 & 50.7 & 30.0 & 49.0 & 60.6 \\
%  ours* & Res-101 & 36 & 152M & 227G & 47.5 & 66.8 & 51.8 & 30.3 & 50.7 & 62.5 \\
%  ours* & Swin-T & 36 & 136M & 167G & 49.6 & 69.4 & 54.4 & 33.8 & 51.9 & 64.4 \\ 
 ours & Res-50 & 36 & 133M & 137G & 44.4 & 63.4 & 48.3 & 27.5 & 47.1 & 58.2 \\
 ours & Res-101 & 36 & 152M & 203G & 45.6 & 64.4 & 49.8 & 27.5 & 48.5 & 60.2 \\
 ours* & Res-50 & 36 & 133M & 160G & 46.4 & 65.8 & 50.7 & 29.9 & 49.0 & 60.5 \\
 ours* & Res-101 & 36 & 152M & 227G & 47.5 & 66.8 & 51.8 & 30.4 & 50.7 & 62.5 \\
 ours* & Swin-T & 36 & 136M & 167G & 49.6 & 69.4 & 54.4 & 33.8 & 51.9 & 64.4 \\
\bottomrule
\end{tabular}
\end{table}

\setlength{\tabcolsep}{5pt}
\begin{table*}[htbp]
\centering
\caption{Performances of different detectors on COCO test-dev set. `ReX' indicates ResNeXt, and `TTA' indicates test-time augmentations, following the settings in \cite{zhang2020bridging}.}
\label{table:res_coco_test}
% \vspace{0.2cm}
\begin{tabular}{c|cc|cccccc}
\toprule
 Method & Backbone & TTA & $\text{AP}$ &  $\text{AP}_{50}$ & $\text{AP}_{75}$ & $\text{AP}_S$ & $\text{AP}_M$ & $\text{AP}_L$ \\ \midrule
 Faster RCNN & Res-101 &   & 36.2 & 59.1 & 39.0 & 18.2 & 39.0  & 48.2 \\
 RetinaNet & Res-101 &   & 39.1 & 59.1 & 42.3 & 21.8 & 42.7 & 50.2 \\
 Cascade R-CNN & Res-101 &   & 42.8 & 62.1 & 46.3 & 23.7 & 45.5  & 55.2 \\
 TSD & Res-101 &   & 43.2 & 64.0 & 46.9 & 24.0 & 46.3 & 55.8 \\
%  Dynamic RCNN & ResNet-101 &   & 44.7 & 63.6 & 49.1 & 26.0 & 47.4 & 57.2 \\
 TSP-RCNN & Res-101 &   & 46.6 & 66.2 & 51.3 & 28.4 & 49.0 & 58.5 \\
%  RepPoint & ResNet-101-DCN &   & 45.0 & 66.1 & 49.0 & 26.6 & 48.6 & 57.5 \\
 Sparse R-CNN* & ReX-101 &   & 46.9 & 66.3 & 51.2 & 28.6 & 49.2 & 58.7 \\
 \midrule
 ours & Res-101 &   & 45.6 & 64.9 & 49.6 & 26.9 & 47.9 & 58.4 \\
 ours* & Res-101 &   & 47.8 & 67.1 & 52.3 & 29.0 & 50.1 & 60.4 \\
%  ours & ResNet-50 & \checkmark & 48.3 & 67.5 & 53.7 & 31.9 & 49.9 & 60.8 \\
 ours & Res-101 & \checkmark & 49.5 & 68.7 & 54.9 & 32.4 & 51.5 & 61.9 \\
%  ours* & ResNet-50 & \checkmark & 49.4 & 68.8 & 55.2 & 32.9 & 51.1 & 61.4 \\
 ours* & Res-101 & \checkmark & 50.8 & 70.2 & 56.8 & 34.1 & 52.7 & 63.1 \\
 ours* & Swin-T & \checkmark & 52.2 & 71.9 & 58.4 & 35.4 & 53.7 & 65.0 \\
\bottomrule
\end{tabular}
\end{table*}

\subsubsection{Main results on COCO.}
\cref{table:res_coco_val} shows the performances of our proposed methods with two standard backbones, ResNet-50 and ResNet-101, and it also compares with some mainstream object detectors or DETR-series. 
It can be seen that the sparse R-CNN outperforms well-established detectors, such as Faster R-CNN and RetinaNet.
Our enhanced version can further improve its detection accuracy. 
Our work reaches 44.4 AP, with 100 query slots based on ResNet-50, which gains 4.2 and 1.6 AP from Faster R-CNN and sparse R-CNN. 
Note that our model only slightly increases the calculations from 134 to 137 GFlops. However, it is still efficient compared with other models like deformable DETR, TSP and SMCA.
With 300 query slots, we can achieve 46.4 on AP, which is not only superior to 45.0 from sparse R-CNN, but also outperforms other methods with even bigger backbone like ResNet-101.
Our method consistently boosts the performance under ResNet-101. Particularly, it gives the AP at 45.6 and 47.5 with 100 and 300 query slots, respectively, compared with 44.1 and 46.4 from sparse R-CNN of the same settings. 
More surprisingly, our method reaches 49.6 on the Swin-T backbone, surpassing sparse R-CNN 1.7 AP.
The results of our method on COCO test-dev set are reported in \cref{table:res_coco_test}. With the help of test-time augmentation, the performance can be further improved. 

%===========================================================
\section{Conclusion}
This paper enhances sparse R-CNN and makes it more suitable for dense object detection. We design two modules, including IoU-ESA and DCW. 
The former strengthens the original self attention performed on object queries. 
By employing IoU of two corresponding proposal boxes, the object query is updated by other relevant ones under the guide from spatial relations. 
The latter allows the object query to provide two extra dynamic channel weights for classification and regression, so image feature within proposal boxes can be highlighted in different ways to meet the requirement of the two detection tasks.  
Both two modules effectively boost the performance of spare R-CNN. 
They can significantly increase the metrics on CrowdHuman and MS-COCO based on different backbones. 
Considering the fact that sparse R-CNN has relatively more parameters than other methods, particularly in its multi-stage dynamic heads, we will explore the ways to reduce model size and to increase the efficiency.

%------------------------------------------------------------------------- 
\clearpage

\bibliographystyle{splncs04}
\bibliography{egbib}

\begin{thebibliography}{10}
\providecommand{\url}[1]{\texttt{#1}}
\providecommand{\urlprefix}{URL }
\providecommand{\doi}[1]{https://doi.org/#1}

\bibitem{bodla2017soft}
Bodla, N., Singh, B., Chellappa, R., Davis, L.S.: Soft-nms--improving object
  detection with one line of code. In: Proceedings of the IEEE international
  conference on computer vision. pp. 5561--5569 (2017)

\bibitem{cai2018cascade}
Cai, Z., Vasconcelos, N.: Cascade r-cnn: Delving into high quality object
  detection. In: Proceedings of the IEEE conference on computer vision and
  pattern recognition. pp. 6154--6162 (2018)

\bibitem{carion2020end}
Carion, N., Massa, F., Synnaeve, G., Usunier, N., Kirillov, A., Zagoruyko, S.:
  End-to-end object detection with transformers. In: European Conference on
  Computer Vision. pp. 213--229. Springer (2020)

\bibitem{chen2019mmdetection}
Chen, K., Wang, J., Pang, J., Cao, Y., Xiong, Y., Li, X., Sun, S., Feng, W.,
  Liu, Z., Xu, J., et~al.: Mmdetection: Open mmlab detection toolbox and
  benchmark. arXiv preprint arXiv:1906.07155  (2019)

\bibitem{chen2020reppoints}
Chen, Y., Zhang, Z., Cao, Y., Wang, L., Lin, S., Hu, H.: Reppoints v2:
  Verification meets regression for object detection. Advances in Neural
  Information Processing Systems  \textbf{33} (2020)

\bibitem{dai2017deformable}
Dai, J., Qi, H., Xiong, Y., Li, Y., Zhang, G., Hu, H., Wei, Y.: Deformable
  convolutional networks. In: Proceedings of the IEEE international conference
  on computer vision. pp. 764--773 (2017)

\bibitem{dalal2005histograms}
Dalal, N., Triggs, B.: Histograms of oriented gradients for human detection.
  In: 2005 IEEE computer society conference on computer vision and pattern
  recognition (CVPR'05). vol.~1, pp. 886--893. Ieee (2005)

\bibitem{deng2009imagenet}
Deng, J., Dong, W., Socher, R., Li, L.J., Li, K., Fei-Fei, L.: Imagenet: A
  large-scale hierarchical image database. In: 2009 IEEE conference on computer
  vision and pattern recognition. pp. 248--255. Ieee (2009)

\bibitem{dollar2011pedestrian}
Dollar, P., Wojek, C., Schiele, B., Perona, P.: Pedestrian detection: An
  evaluation of the state of the art. IEEE transactions on pattern analysis and
  machine intelligence  \textbf{34}(4),  743--761 (2011)

\bibitem{dosovitskiy2020image}
Dosovitskiy, A., Beyer, L., Kolesnikov, A., Weissenborn, D., Zhai, X.,
  Unterthiner, T., Dehghani, M., Minderer, M., Heigold, G., Gelly, S., et~al.:
  An image is worth 16x16 words: Transformers for image recognition at scale.
  arXiv preprint arXiv:2010.11929  (2020)

\bibitem{gao2021fast}
Gao, P., Zheng, M., Wang, X., Dai, J., Li, H.: Fast convergence of detr with
  spatially modulated co-attention. arXiv preprint arXiv:2101.07448  (2021)

\bibitem{ge2021ota}
Ge, Z., Liu, S., Li, Z., Yoshie, O., Sun, J.: Ota: Optimal transport assignment
  for object detection. In: Proceedings of the IEEE/CVF Conference on Computer
  Vision and Pattern Recognition. pp. 303--312 (2021)

\bibitem{girshick2015fast}
Girshick, R.: Fast r-cnn. In: Proceedings of the IEEE international conference
  on computer vision. pp. 1440--1448 (2015)

\bibitem{girshick2014rich}
Girshick, R., Donahue, J., Darrell, T., Malik, J.: Rich feature hierarchies for
  accurate object detection and semantic segmentation. In: Proceedings of the
  IEEE conference on computer vision and pattern recognition. pp. 580--587
  (2014)

\bibitem{Detectron2018}
Girshick, R., Radosavovic, I., Gkioxari, G., Doll\'{a}r, P., He, K.: Detectron.
  \url{https://github.com/facebookresearch/detectron} (2018)

\bibitem{glorot2010understanding}
Glorot, X., Bengio, Y.: Understanding the difficulty of training deep
  feedforward neural networks. In: Proceedings of the thirteenth international
  conference on artificial intelligence and statistics. pp. 249--256. JMLR
  Workshop and Conference Proceedings (2010)

\bibitem{he2017mask}
He, K., Gkioxari, G., Doll{\'a}r, P., Girshick, R.: Mask r-cnn. In: Proceedings
  of the IEEE international conference on computer vision. pp. 2961--2969
  (2017)

\bibitem{he2016deep}
He, K., Zhang, X., Ren, S., Sun, J.: Deep residual learning for image
  recognition. In: Proceedings of the IEEE conference on computer vision and
  pattern recognition. pp. 770--778 (2016)

\bibitem{hu2018relation}
Hu, H., Gu, J., Zhang, Z., Dai, J., Wei, Y.: Relation networks for object
  detection. In: Proceedings of the IEEE conference on computer vision and
  pattern recognition. pp. 3588--3597 (2018)

\bibitem{hu2018squeeze}
Hu, J., Shen, L., Sun, G.: Squeeze-and-excitation networks. In: Proceedings of
  the IEEE conference on computer vision and pattern recognition. pp.
  7132--7141 (2018)

\bibitem{huang2020nms}
Huang, X., Ge, Z., Jie, Z., Yoshie, O.: Nms by representative region: Towards
  crowded pedestrian detection by proposal pairing. In: Proceedings of the
  IEEE/CVF Conference on Computer Vision and Pattern Recognition. pp.
  10750--10759 (2020)

\bibitem{kim2020probabilistic}
Kim, K., Lee, H.S.: Probabilistic anchor assignment with iou prediction for
  object detection. In: Computer Vision--ECCV 2020: 16th European Conference,
  Glasgow, UK, August 23--28, 2020, Proceedings, Part XXV 16. pp. 355--371.
  Springer (2020)

\bibitem{law2018cornernet}
Law, H., Deng, J.: Cornernet: Detecting objects as paired keypoints. In:
  Proceedings of the European conference on computer vision (ECCV). pp.
  734--750 (2018)

\bibitem{lin2017feature}
Lin, T.Y., Doll{\'a}r, P., Girshick, R., He, K., Hariharan, B., Belongie, S.:
  Feature pyramid networks for object detection. In: Proceedings of the IEEE
  conference on computer vision and pattern recognition. pp. 2117--2125 (2017)

\bibitem{lin2017focal}
Lin, T.Y., Goyal, P., Girshick, R., He, K., Doll{\'a}r, P.: Focal loss for
  dense object detection. In: Proceedings of the IEEE international conference
  on computer vision. pp. 2980--2988 (2017)

\bibitem{lin2014microsoft}
Lin, T.Y., Maire, M., Belongie, S., Hays, J., Perona, P., Ramanan, D.,
  Doll{\'a}r, P., Zitnick, C.L.: Microsoft coco: Common objects in context. In:
  European conference on computer vision. pp. 740--755. Springer (2014)

\bibitem{liu2019adaptive}
Liu, S., Huang, D., Wang, Y.: Adaptive nms: Refining pedestrian detection in a
  crowd. In: Proceedings of the IEEE/CVF Conference on Computer Vision and
  Pattern Recognition. pp. 6459--6468 (2019)

\bibitem{liu2016ssd}
Liu, W., Anguelov, D., Erhan, D., Szegedy, C., Reed, S., Fu, C.Y., Berg, A.C.:
  Ssd: Single shot multibox detector. In: European conference on computer
  vision. pp. 21--37. Springer (2016)

\bibitem{liu2021swin}
Liu, Z., Lin, Y., Cao, Y., Hu, H., Wei, Y., Zhang, Z., Lin, S., Guo, B.: Swin
  transformer: Hierarchical vision transformer using shifted windows. arXiv
  preprint arXiv:2103.14030  (2021)

\bibitem{loshchilov2018decoupled}
Loshchilov, I., Hutter, F.: Decoupled weight decay regularization. In:
  International Conference on Learning Representations (2018)

\bibitem{redmon2016you}
Redmon, J., Divvala, S., Girshick, R., Farhadi, A.: You only look once:
  Unified, real-time object detection. In: Proceedings of the IEEE conference
  on computer vision and pattern recognition. pp. 779--788 (2016)

\bibitem{redmon2017yolo9000}
Redmon, J., Farhadi, A.: Yolo9000: better, faster, stronger. In: Proceedings of
  the IEEE conference on computer vision and pattern recognition. pp.
  7263--7271 (2017)

\bibitem{ren2015faster}
Ren, S., He, K., Girshick, R., Sun, J.: Faster r-cnn: Towards real-time object
  detection with region proposal networks. Advances in neural information
  processing systems  \textbf{28},  91--99 (2015)

\bibitem{rezatofighi2019generalized}
Rezatofighi, H., Tsoi, N., Gwak, J., Sadeghian, A., Reid, I., Savarese, S.:
  Generalized intersection over union: A metric and a loss for bounding box
  regression. In: Proceedings of the IEEE/CVF Conference on Computer Vision and
  Pattern Recognition. pp. 658--666 (2019)

\bibitem{shao2018crowdhuman}
Shao, S., Zhao, Z., Li, B., Xiao, T., Yu, G., Zhang, X., Sun, J.: Crowdhuman: A
  benchmark for detecting human in a crowd. arXiv preprint arXiv:1805.00123
  (2018)

\bibitem{song2020revisiting}
Song, G., Liu, Y., Wang, X.: Revisiting the sibling head in object detector.
  In: Proceedings of the IEEE/CVF Conference on Computer Vision and Pattern
  Recognition. pp. 11563--11572 (2020)

\bibitem{stewart2016end}
Stewart, R., Andriluka, M., Ng, A.Y.: End-to-end people detection in crowded
  scenes. In: Proceedings of the IEEE conference on computer vision and pattern
  recognition. pp. 2325--2333 (2016)

\bibitem{sun2020onenet}
Sun, P., Jiang, Y., Xie, E., Yuan, Z., Wang, C., Luo, P.: Onenet: Towards
  end-to-end one-stage object detection. arXiv preprint arXiv:2012.05780
  (2020)

\bibitem{sun2021sparse}
Sun, P., Zhang, R., Jiang, Y., Kong, T., Xu, C., Zhan, W., Tomizuka, M., Li,
  L., Yuan, Z., Wang, C., et~al.: Sparse r-cnn: End-to-end object detection
  with learnable proposals. In: Proceedings of the IEEE/CVF Conference on
  Computer Vision and Pattern Recognition. pp. 14454--14463 (2021)

\bibitem{sun2020rethinking}
Sun, Z., Cao, S., Yang, Y., Kitani, K.: Rethinking transformer-based set
  prediction for object detection. arXiv preprint arXiv:2011.10881  (2020)

\bibitem{tian2019fcos}
Tian, Z., Shen, C., Chen, H., He, T.: Fcos: Fully convolutional one-stage
  object detection. In: Proceedings of the IEEE/CVF international conference on
  computer vision. pp. 9627--9636 (2019)

\bibitem{vaswani2017attention}
Vaswani, A., Shazeer, N., Parmar, N., Uszkoreit, J., Jones, L., Gomez, A.N.,
  Kaiser, {\L}., Polosukhin, I.: Attention is all you need. In: Advances in
  neural information processing systems. pp. 5998--6008 (2017)

\bibitem{viola2001rapid}
Viola, P., Jones, M.: Rapid object detection using a boosted cascade of simple
  features. In: Proceedings of the 2001 IEEE computer society conference on
  computer vision and pattern recognition. CVPR 2001. vol.~1, pp.~I--I. Ieee
  (2001)

\bibitem{wang2021pyramid}
Wang, W., Xie, E., Li, X., Fan, D.P., Song, K., Liang, D., Lu, T., Luo, P.,
  Shao, L.: Pyramid vision transformer: A versatile backbone for dense
  prediction without convolutions. arXiv preprint arXiv:2102.12122  (2021)

\bibitem{yang2019reppoints}
Yang, Z., Liu, S., Hu, H., Wang, L., Lin, S.: Reppoints: Point set
  representation for object detection. In: Proceedings of the IEEE/CVF
  International Conference on Computer Vision. pp. 9657--9666 (2019)

\bibitem{zhang2020bridging}
Zhang, S., Chi, C., Yao, Y., Lei, Z., Li, S.Z.: Bridging the gap between
  anchor-based and anchor-free detection via adaptive training sample
  selection. In: Proceedings of the IEEE/CVF conference on computer vision and
  pattern recognition. pp. 9759--9768 (2020)

\bibitem{zhao2019object}
Zhao, Z.Q., Zheng, P., Xu, S.t., Wu, X.: Object detection with deep learning: A
  review. IEEE transactions on neural networks and learning systems
  \textbf{30}(11),  3212--3232 (2019)

\bibitem{zhou2019objects}
Zhou, X., Wang, D., Kr{\"a}henb{\"u}hl, P.: Objects as points. arXiv preprint
  arXiv:1904.07850  (2019)

\bibitem{zhu2019deformable}
Zhu, X., Hu, H., Lin, S., Dai, J.: Deformable convnets v2: More deformable,
  better results. In: Proceedings of the IEEE/CVF Conference on Computer Vision
  and Pattern Recognition. pp. 9308--9316 (2019)

\bibitem{zhu2020deformable}
Zhu, X., Su, W., Lu, L., Li, B., Wang, X., Dai, J.: Deformable detr: Deformable
  transformers for end-to-end object detection. In: International Conference on
  Learning Representations (2020)

\end{thebibliography}

%this would normally be the end of your paper, but you may also have an appendix
%within the given limit of number of pages

\clearpage
\appendix
{\LARGE\noindent\textbf{Appendix}}

\section{Further Illustrations and Pseudo Codes}
Here, we describe the implementations of the two designed modules in detail.  
Particularly, IoU-ESA and DCW are summarized into %the 
PyTorch-like code in \cref{alg:IoU-ESA} and \cref{alg:DCW}, respectively. 
The complete code is provided in the file of the supplementary material.

For IoU-ESA, it is defined in Eq.(3) in the paper, without newly added parameters.
The key difference with the original self attention is that the $IoU$ matrix calculated among proposal boxes, is element-wise multiplied with the attention matrix. 
Both the $IoU$ and $attn$ are reflecting a kind of similarity of different proposal features.

For the DCW module, two light weights projection heads are added to generate two channel masks for classification and localization, respectively. %regression. 
Because of the the two fc bottlenecks, the increase of parameters and calculations is also small. 
This module strengthens the two head features for a better utilization of object queries. It can be seen in Fig.1 in the paper.

\begin{algorithm}[ht]
\caption{{IoU-ESA code (PyTorch-like)}}
\label{alg:IoU-ESA}
\definecolor{codeblue}{rgb}{0.25,0.5,0.25}
\lstset{
    language=python,
	backgroundcolor=\color{white},
	basicstyle=\fontsize{8pt}{8pt}\ttfamily\selectfont,
	columns=fullflexible,
	breaklines=true,
	captionpos=b,
	commentstyle=\fontsize{8pt}{8pt}\color{codeblue},
	keywordstyle=\fontsize{8pt}{8pt}, %\color{blue},
% 	morekeywords={torch,exp},
}
% \begin{minted}[fontsize=\scriptsize]{python}
\begin{lstlisting}[language=python]
# x: input tensor (N, d); 
# iou: (N, N)

def IoU-ESA(x, iou):
    q, k, v = fc_qkv(x).chunk(3, dim=-1)
    # nhead: num of heads in attention
    q = q.view(N, nhead, -1).transpose(0, 1)
    k = k.view(N, nhead, -1).transpose(0, 1)
    v = v.view(N, nhead, -1).transpose(0, 1)
    
    attn = torch.bmm(q, k.transpose(1, 2))
    
    # Eq (3). in the paper
    attn = torch.exp(attn)
    attn = attn * iou[None, :, :]
    attn = attn / torch.sum(attn, -1, keepdim=True)
    
    # value routing and output projection
    attn = torch.bmm(attn, v)
    attn= attn.transpose(0, 1).view(N, d)
    attn_out = fc_out(attn)

    return attn_out
\end{lstlisting}
\end{algorithm}

\begin{algorithm}[ht]
\caption{{DCW code (PyTorch-like)}}
\label{alg:DCW}
\definecolor{codeblue}{rgb}{0.25,0.5,0.25}
\lstset{
	backgroundcolor=\color{white},
	basicstyle=\fontsize{8pt}{8pt}\ttfamily\selectfont,
	columns=fullflexible,
	breaklines=true,
	captionpos=b,
	commentstyle=\fontsize{8pt}{8pt}\color{codeblue},
	keywordstyle=\fontsize{8pt}{8pt}%\color{blue}
}
% \begin{minted}[fontsize=\scriptsize]{python}
\begin{lstlisting}[language=python]
# obj_queries: (N, d)
# roi_feats: roi features, output of dynamic convs (N, s*s, d)

def DCW(pro_feats, roi_feats):
    # Eq.(4). in the paper
    # mask: (N, d)
    mask_cls = sigmoid(fc2(fc1(obj_queries)))
    mask_reg = sigmoid(fc4(fc3(obj_queries)))
    
    # roi_feats: (N, s*s, d)
    roi_feats_cls = roi_feats[:, None, :] * mask_cls
    roi_feats_reg = roi_feats[:, None, :] * mask_reg
    
    # roi_feats: (N, s*s*d)
    roi_feats_cls = roi_feats_cls.flatten(1)
    roi_feats_reg = roi_feats_reg.flatten(1)
    
    # obj_feats: (N, d)
    obj_feats_cls = fc5(roi_feats_cls)
    obj_feats_reg = fc6(roi_feats_reg)

    return obj_feats_cls, obj_feats_reg
\end{lstlisting}
\end{algorithm}

\section{Analyze result on CrowdHuman}
 We analyze self attention and dimension of features in Sec.3.1 in the paper. Results on MS-COCO illustrate that geometry prior is expected to guide self attention and disentangle features for the two tasks in detection is effective. Here, we provide the the results on CrowdHuman in \cref{table:analyze_on_selfattn_crowdhuman} and \cref{table:analyze_on_feature_dim_crowdhuman}.
 Performance on CrowdHuman also illustrate our motivation.

\begin{table}[ht]
\centering
\caption{Analysis on self attention. 
Three different models, including origin sparse R-CNN with multi-head self attention (MSA), 
without self attention and attention matrix replaced by IoU matrix (IoU-MSA), are trained on CrowdHuman. 
AP and mMR indicate the Average Precision and Log-average Miss Rate, respectively.}
% \vspace{0.2cm}
\begin{tabular}{@{}cccc@{}}
\toprule
Method & AP $\uparrow$ &  mMR $\downarrow$ & Recall $\uparrow$ \\ \midrule
MSA & 88.8 & 49.9 & 95.8 \\
w/o MSA & 81.5 & 65.5 & 95.5 \\
IoU-MSA & 86.8 & 54.2 & 95.7 \\ \bottomrule
\end{tabular}
\label{table:analyze_on_selfattn_crowdhuman}
\end{table}

\begin{table}[ht]
\centering
\caption{Analysis on feature disentanglement on CrowdHuman.
}
% \vspace{0.2cm}
\begin{tabular}{@{}cccc@{}}
\toprule
Method & AP $\uparrow$ &  mMR $\downarrow$ & Recall $\uparrow$ \\ \midrule
En. (original) & 89.2 & 48.3 &  95.9 \\
Dis. (half dim) & 88.6 & 49.7 & 95.5 \\
Dis. (full dim) & 89.1 & 48.5 & 95.8 \\ \bottomrule
\end{tabular}
\label{table:analyze_on_feature_dim_crowdhuman}
\end{table}

% \section{Discussions and Limitations}
\section{Implementation on DETR and discussions}
We also insert the two modules, IoU-ESA and DCW, into DETR \cite{carion2020end}. DETR and sparse R-CNN \cite{sun2021sparse} both are of cascade structures, and updating object queries for the next stage.
In sparse R-CNN, proposal boxes are inputs of the model, which are parameters at the initial stage and output bounding boxes from last stage at later.
$IoU$ matrix is computed among proposal boxes.
However, there are no bounding boxes as input in DETR. Therefore, $IoU$ is set to 1 for all elements at the first stage, and in the later stage, it is computed in the same as sparse R-CNN. %at the later. 
The channel masks $m_c$ and $m_r$, given by DCW, are generated from the object queries before cross attention, and then perform on object queries after cross attention.

\begin{table*}[htbp]
\centering
\caption{Implementation results on DETR. ResNet-50 is utilized as backbone, and the number of object queries is set as 100.}
% \vspace{0.2cm}
\begin{tabular}{c|cc|cccccc}
\toprule
 Method & Params & FLOPs & $\text{AP}$ &  $\text{AP}_{50}$ & $\text{AP}_{75}$ & $\text{AP}_S$ & $\text{AP}_M$ & $\text{AP}_L$ \\ \midrule
 DETR & 41M & 86G & 34.9 & 55.5 & 35.9 & 14.5 & 37.6 & 53.8 \\
 DETR + ours & 48M & 86G & 35.4 & 56.8 & 36.7 & 15.4 & 38.0 & 54.1\\
\bottomrule
\end{tabular}
\label{tab:detr}
\end{table*}

We train it for 50 epochs, and the learning rate is dropped by a factor of 10 at 40th epoch. All other settings are the same as DETR. As shown in \cref{tab:detr},  our work reaches 35.4, which slightly gains 0.5 AP from DETR. 
It also behaves well on the other five metrics.
The number of parameters in our work rises from 41M to 48M, and FLOPS keeps the same.
Although our work improves the performance on sparse R-CNN significantly, it doesn't do so well on DETR. We notice that quries in DETR obtain all the information from the whole image, which is different from sparse R-CNN. Furthermore, since each query is $1\times 1 \times C$ without spatial dimension, $m_c$ and $m_r$ can not apply for different spatial positions like SE-Net \cite{hu2018squeeze}. 
This indicates that the proposed modules are targeted to sparse R-CNN, in which RoI-align is performed to get the local object region. %IoU is a good metric for evaluating the similarity of two bounding boxes, but .

\section{Visualizations}
\begin{figure}[ht]
    \centering
    \includegraphics[width=0.65\columnwidth]{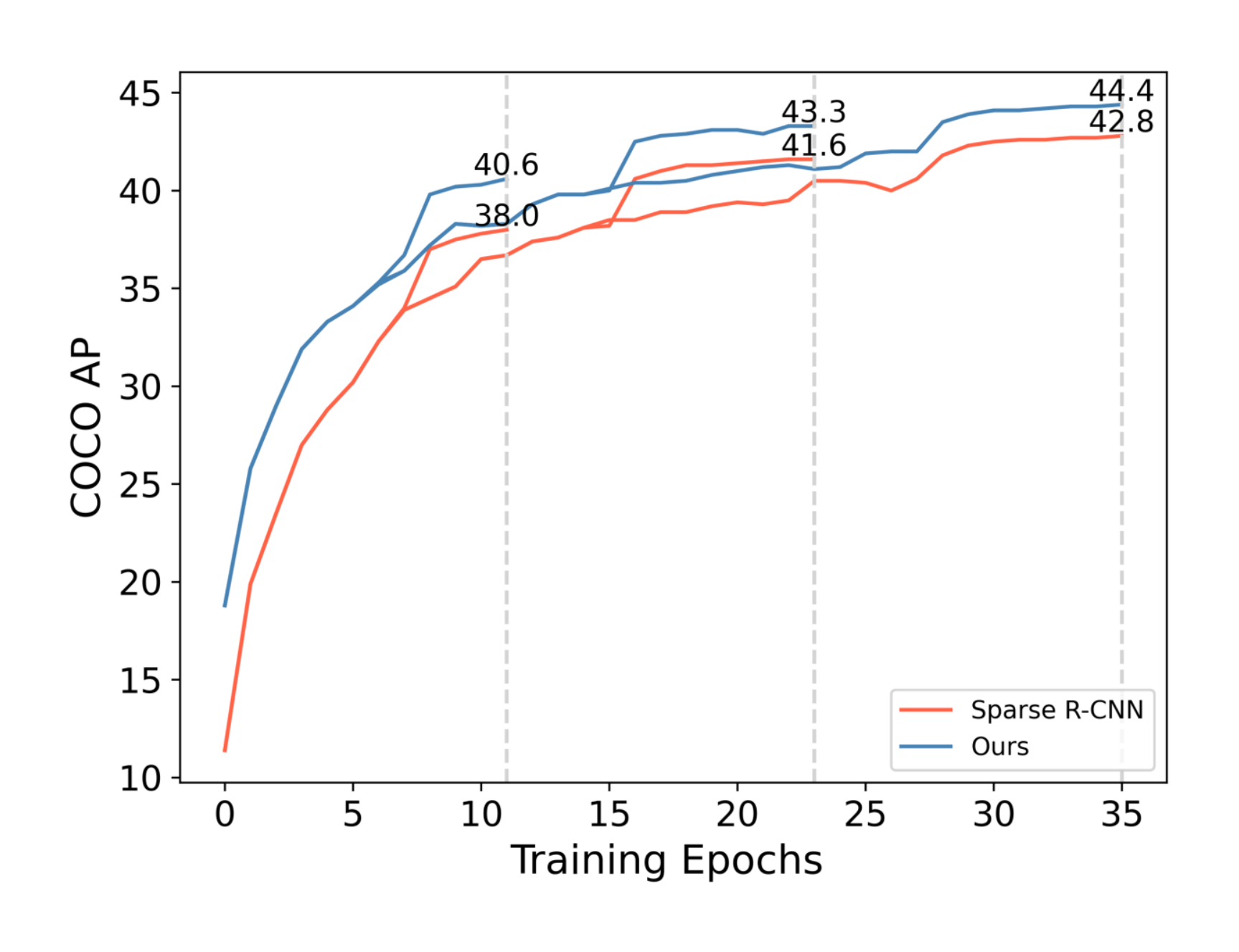}
    \caption{Convergence curves of sparse R-CNN and ours on COCO val2017. %Their 
    Both models have the %training setting is 
    same training settings with ResNet-50 and 100 proposals. Our work gains higher AP than sparse R-CNN from the beginning to the end. %of training.
    }
    \label{fig:convergence}
\end{figure}

\begin{figure*}[ht]
    \centering
    \includegraphics[width=0.95\textwidth]{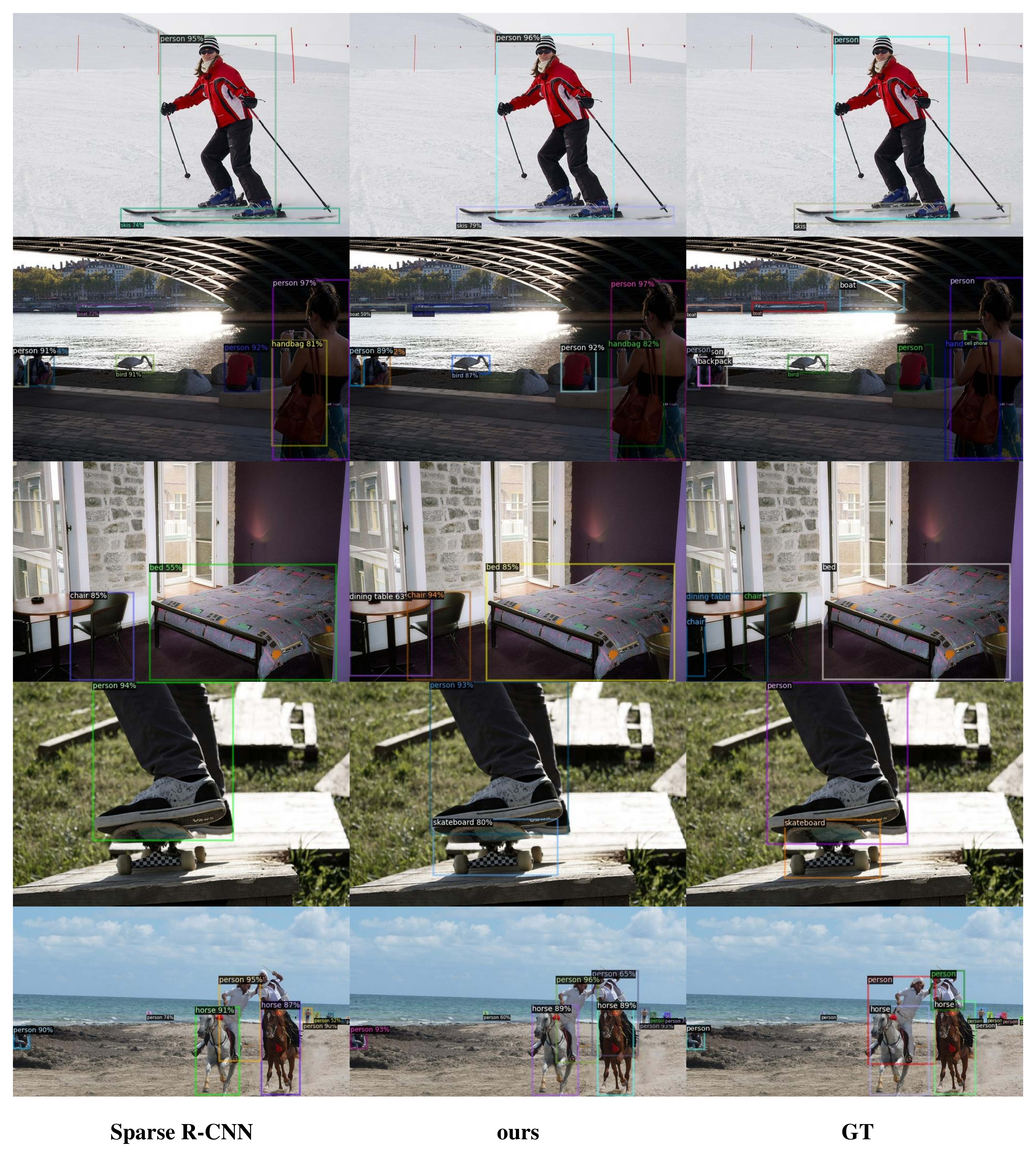}
    \caption{Visualizations of detection results on COCO from sparse R-CNN, ours and ground truths. Our work is more better in complex scenarios.}
    \label{fig:results_coco}
\end{figure*}

\begin{figure*}[ht]
    \centering
    \includegraphics[width=0.95\textwidth]{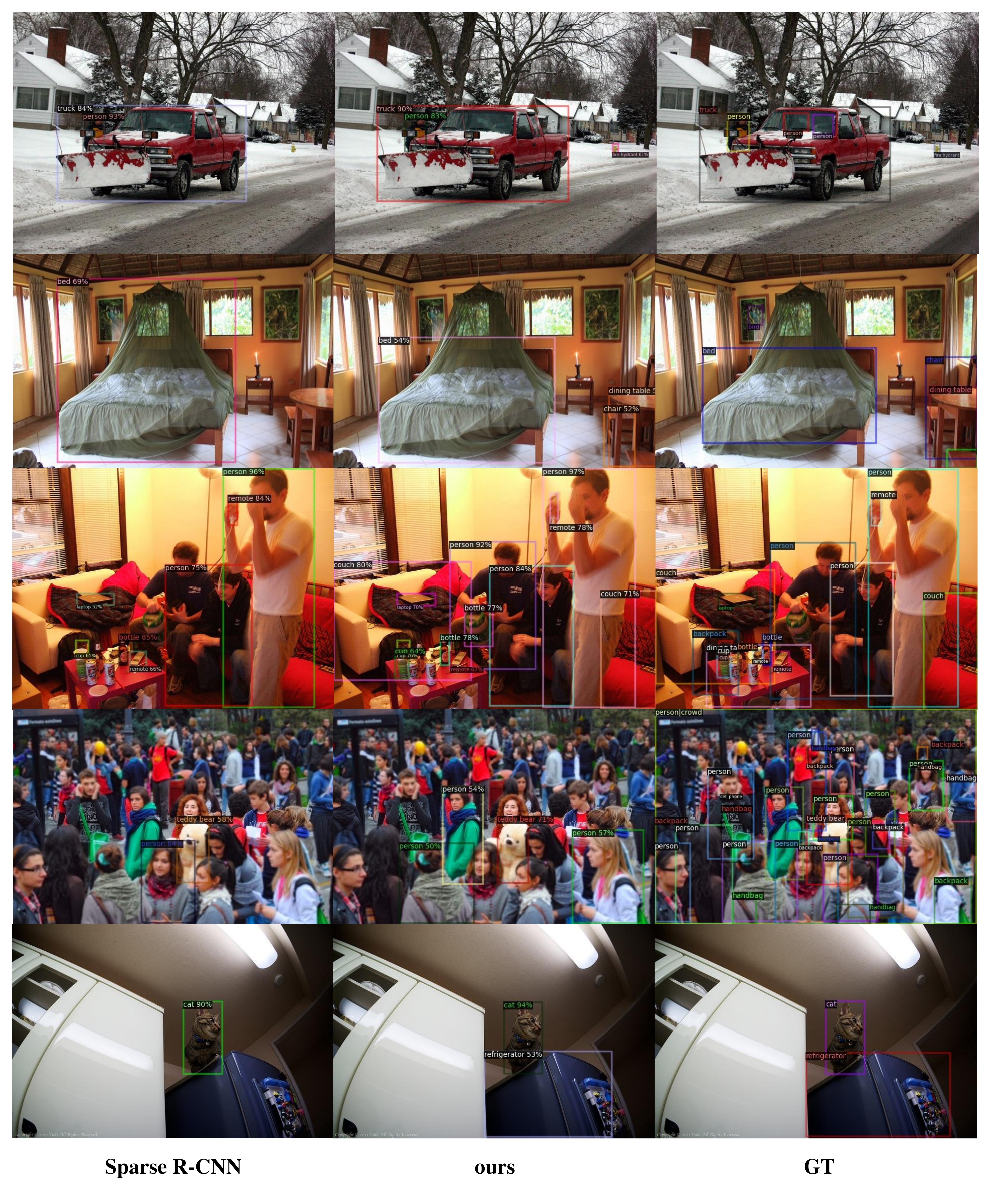}
    \caption{Visualizations of detection results on COCO from sparse R-CNN, ours and ground truths.}
    \label{fig:results_coco2}
\end{figure*}

\begin{figure*}[ht]
    \centering
    \includegraphics[width=0.95\textwidth]{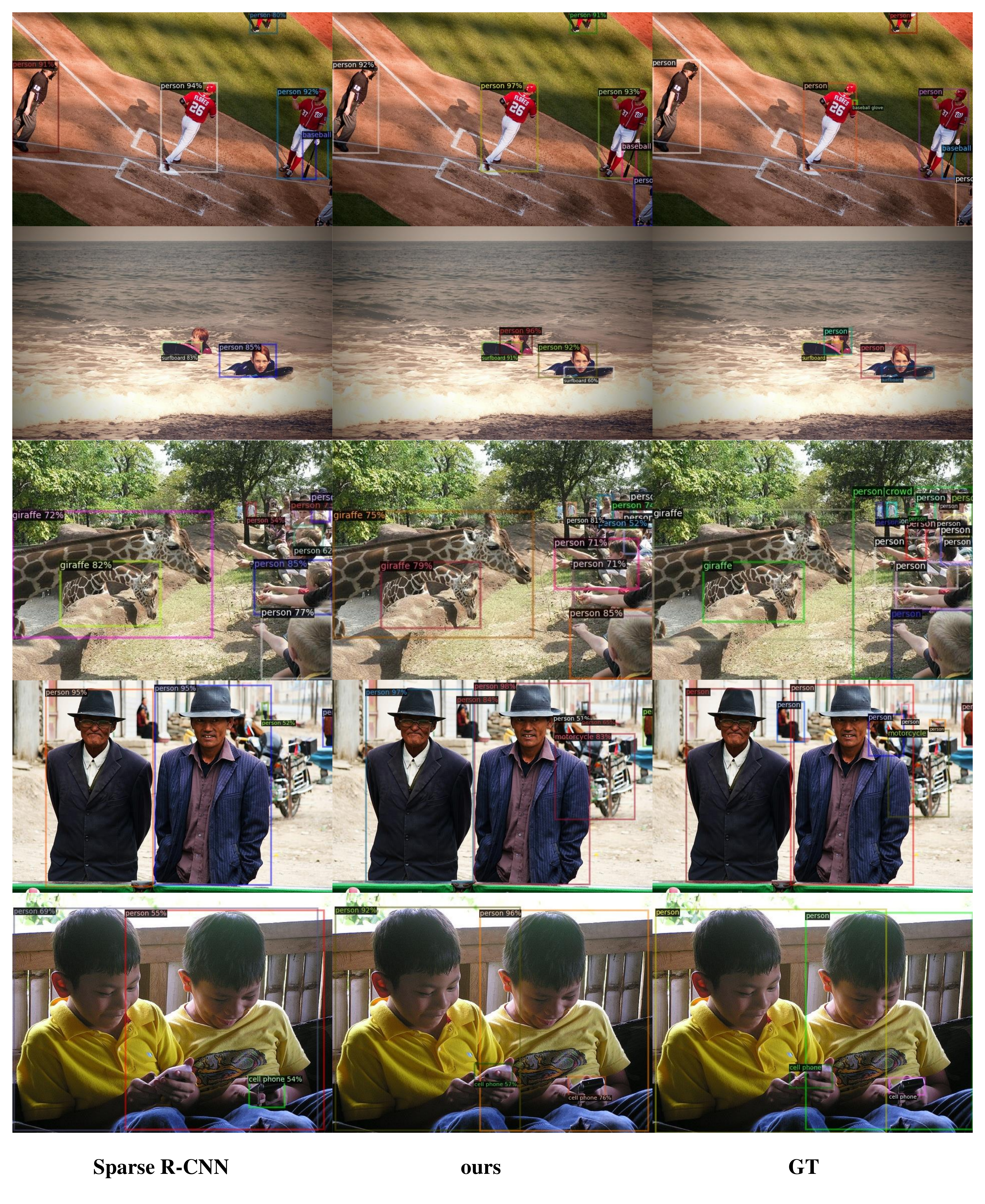}
    \caption{Visualizations of detection results on COCO from sparse R-CNN, ours and ground truths.}
    \label{fig:results_coco3}
\end{figure*}

\begin{figure*}[ht]
    \centering
    \includegraphics[width=0.95\textwidth]{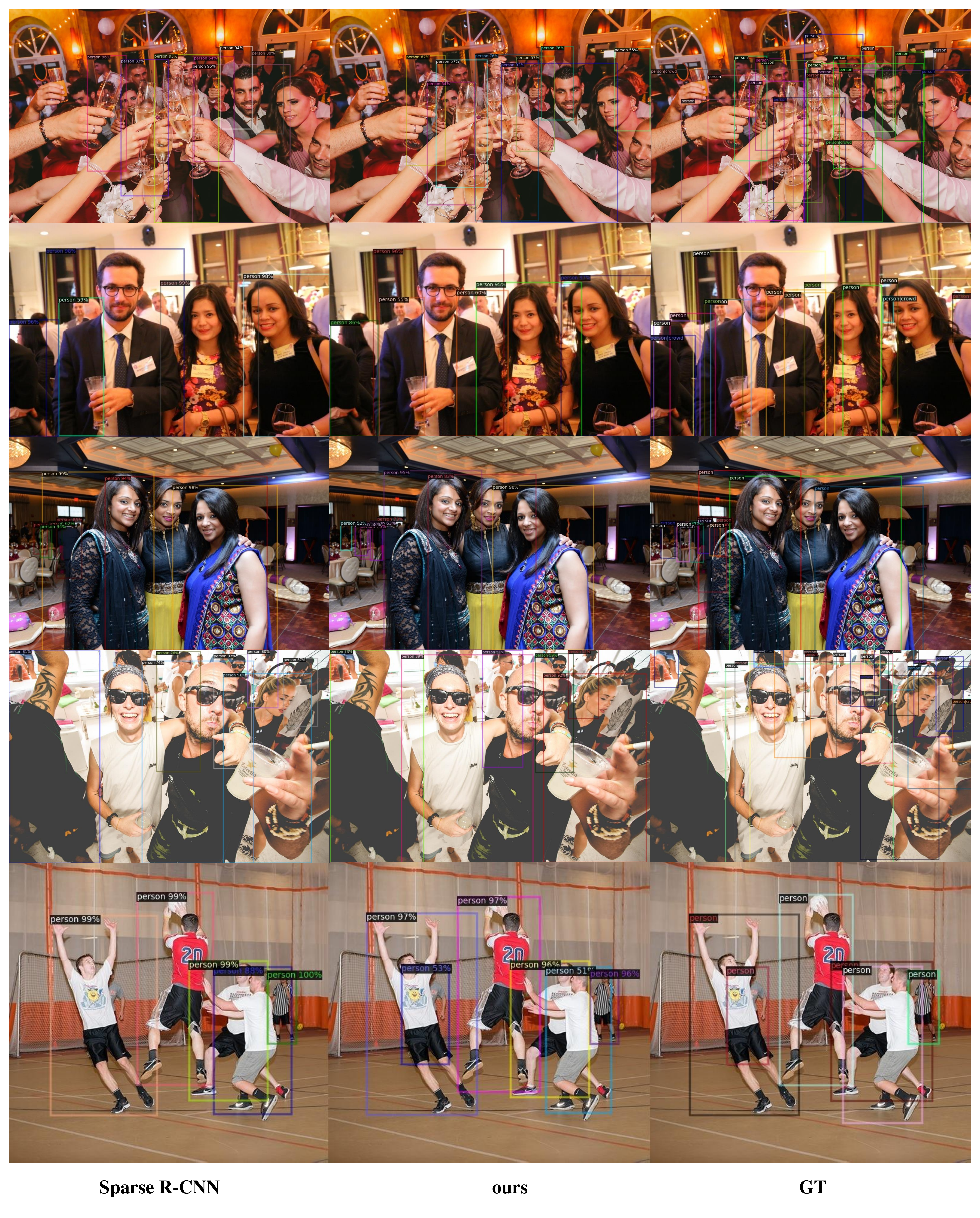}
    \caption{Visualizations of detection results on CrowdHuman from sparse R-CNN, ours and ground truths.}
    \label{fig:results_crowdhuman}
\end{figure*}

As our two designed modules %are using for 
enhance the origin sparse R-CNN, the performance increases stably based on different backbones.
The convergence speed of the two works are shown in \cref{fig:convergence}. 
Note that $1\times$ and $2\times$ (12 and 24 epochs) training schedules are also plotted to indicate the convergence rate.
Compared with sparse R-CNN, our work achieves a better AP from the beginning of training, our work finally gains 1.6 AP at 36th epoch.
Some %detection 
visual results on COCO are giving in \cref{fig:results_coco}, which organizes the three columns in the sequence of sparse R-CNN, ours and ground truths.
In simple scenarios, there are no big differences between sparse R-CNN and our work.
At the first row, both sparse R-CNN and our work have done a good job.
The rest four rows' scenarios are more complex than the first one, we can see that the detection result of our work is better than sparse R-CNN.
The targets in complicated scene are detected by our work. Similar phenomena can be observed in \cref{fig:results_coco2} and \cref{fig:results_coco3}.
Detection results on CrowdHuman are also shown in \cref{fig:results_crowdhuman}.

\end{document}